\documentclass[10pt,twocolumn,letterpaper]{article}

\usepackage{cvpr}
\usepackage{times}
\usepackage{epsfig}
\usepackage{graphicx}
\usepackage{amsmath}
\usepackage{amssymb}

\usepackage[pagebackref=true,breaklinks=true,letterpaper=true,colorlinks,bookmarks=false]{hyperref}

\cvprfinalcopy 

\def\cvprPaperID{800} 

\ifcvprfinal\fi
\begin{document}

\title{Efficient Upsampling of Natural Images}

\author{Chinmay Hegde\thanks{e-mail: chinmay@rice.edu} \\ Rice University
\and Oncel Tuzel\thanks{e-mail:oncel@merl.com} \\ MERL %
\and Fatih Porikli\thanks{e-mail:fatih@merl.com}\\ MERL
}

\maketitle

\begin{abstract}

We propose a novel method of efficient upsampling of a single natural image. Current methods for image upsampling tend to produce high-resolution images with either blurry salient edges, or loss of fine textural detail, or spurious noise artifacts.

In our method, we mitigate these effects by modeling the input image as a sum of edge and detail layers, operating upon these layers separately, and merging the upscaled results in an automatic fashion. We formulate the upsampled output image as the solution to a non-convex energy minimization problem, and propose an algorithm to obtain a tractable approximate solution. Our algorithm comprises two main stages. 1) For the edge layer, we use a {\em nonparametric} approach by constructing a dictionary of patches from a given image, and synthesize edge regions in a higher-resolution version of the image. 2) For the detail layer, we use a global {\em parametric} texture enhancement approach to synthesize detail regions across the image.

We demonstrate that our method is able to accurately reproduce sharp edges as well as synthesize photorealistic textures, while avoiding common artifacts such as ringing and haloing. In addition, our method involves no training phase or estimation of model parameters, and is easily parallelizable. We demonstrate the utility of our method on a number of challenging standard test photos.

\end{abstract}

\section{Introduction}
\label{sec:intro}

\begin{figure}[t]
\begin{tabular}{cc}
\centering
   \includegraphics[height=1.5in]{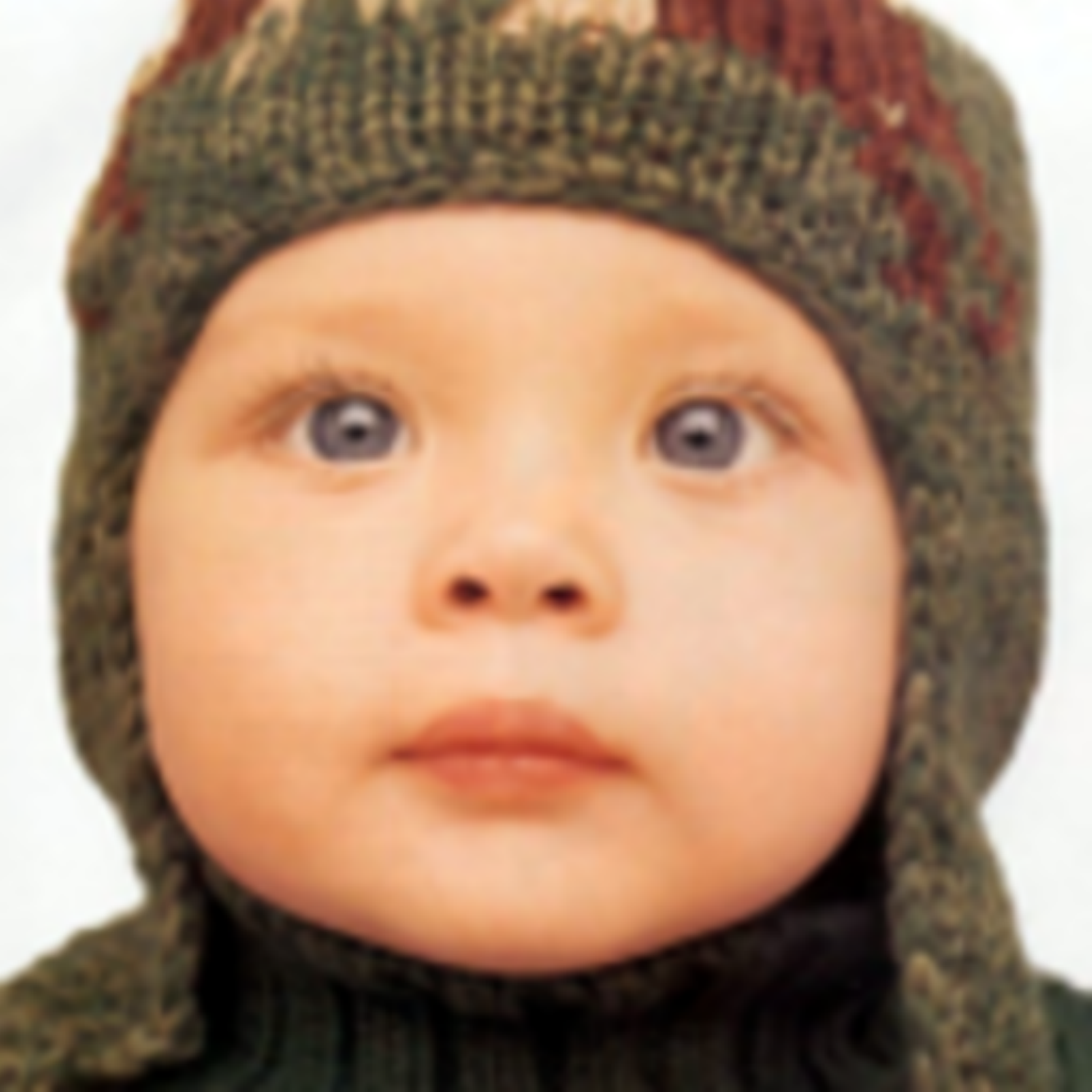} &
   \includegraphics[height=1.5in]{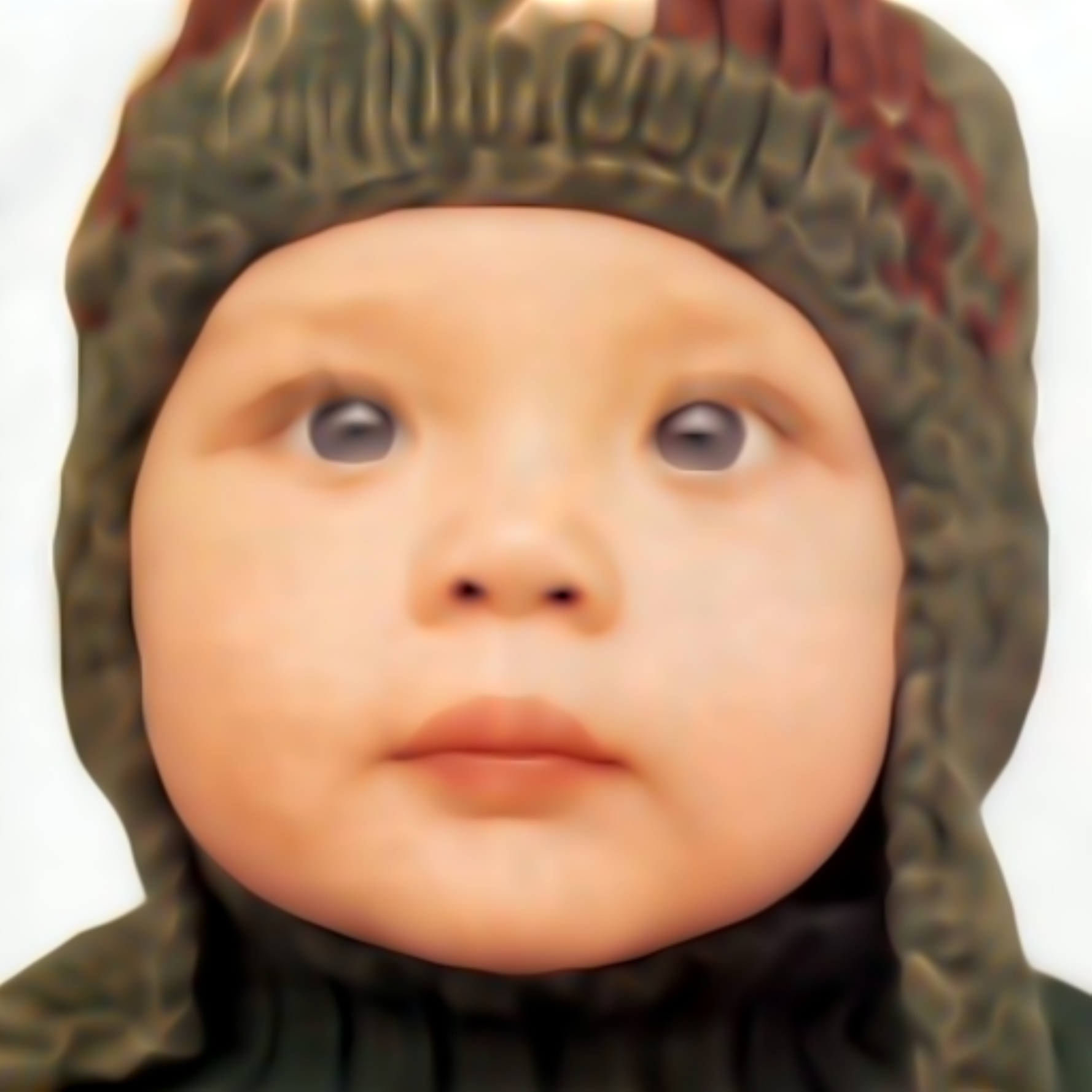} \\
   \includegraphics[height=1.5in]{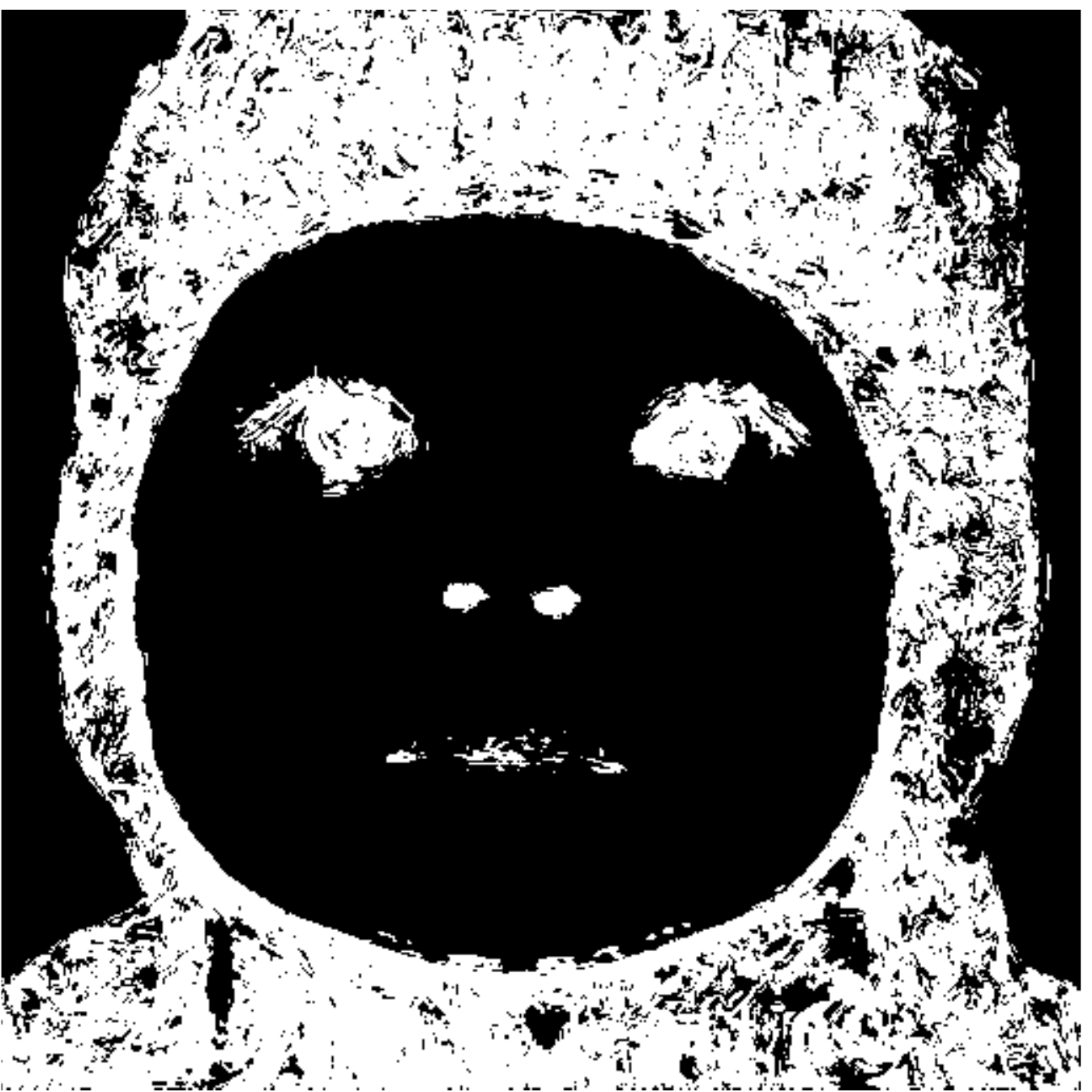} &
   \includegraphics[height=1.5in]{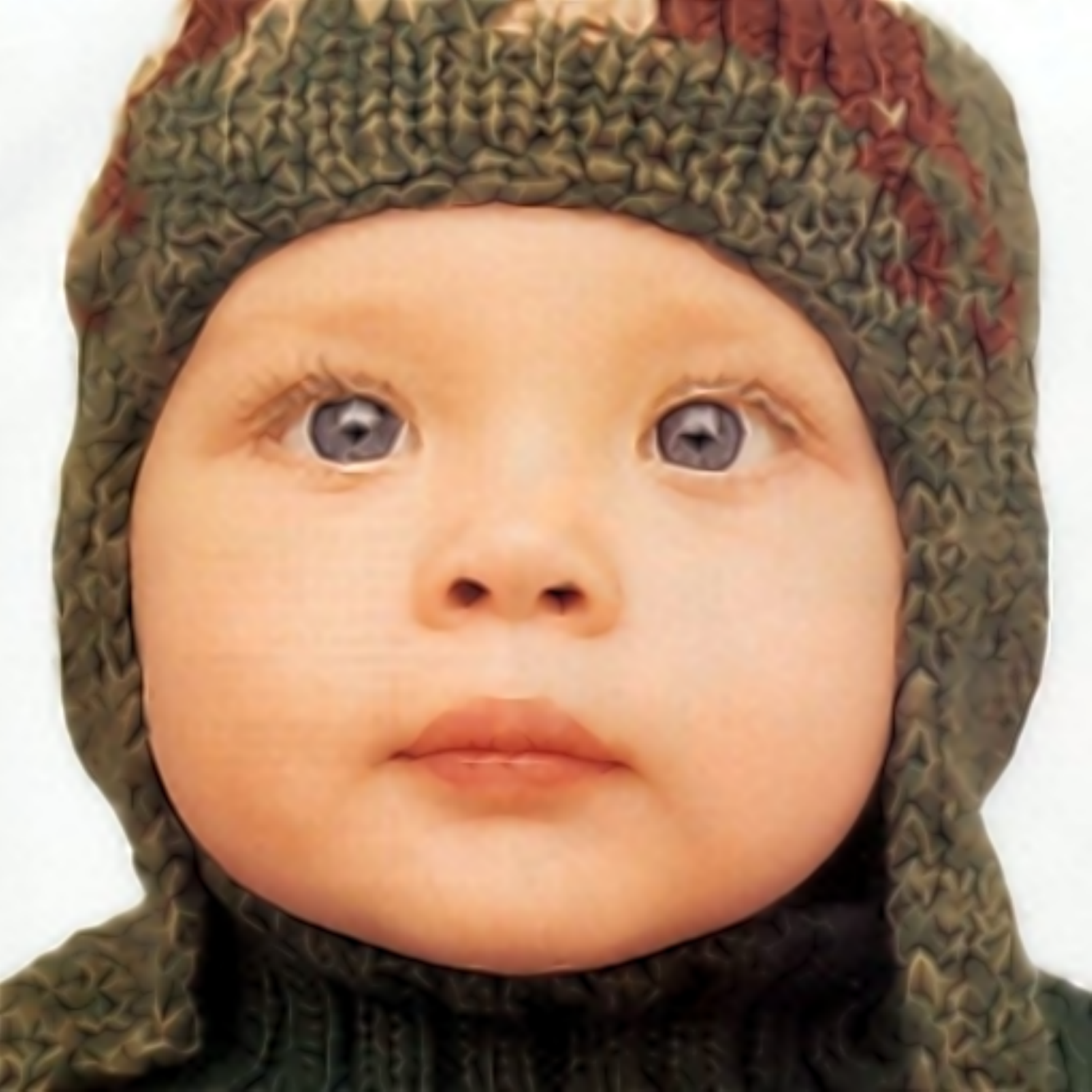}
\end{tabular}
   \caption{Various stages in our upsampling framework. Bicubic interpolation creates an initial estimate $I_1$ (top-left), which is further refined using a localized nonparametric patch synthesis to obtain a second estimate $I_2$ (top-right). While this estimate reproduces smooth regions and sharp edges, it suffers from a noticeable lack of detail. Therefore, it is decomposed into an edge and detail components, and an $S$-shaped curve is applied to achieve a global detail enhancement. To prevent exaggeration of noise and other artifacts, we use a blending $\alpha$-mask (bottom-left), calculated during the synthesis step, to selectively enhance only the detail layer and thus obtain the final image $I$ (bottom-right).}
\label{fig:analysis}
\end{figure}

The last few years in imaging technology has witnessed the advent of ubiquitous high-definition (HD) image display systems, as well as an exponential growth in low-resolution image acquisition sytems, such as cellphone cameras. Given this scenario, the classical problem of {\em image upsampling} assumes a renewed significance.
Image upsampling (also variously termed as {\em upscaling} and {\em super-resolution}) is a highly ill-posed linear inverse problem, since the number of unknowns (high-resolution image pixel values) exceeds the number of observations (low-resolution pixel values) by an order of magnitude, even at moderate upscaling factors. The challenge is exacerbated by the unique difficulties encountered in modeling real-world photos, as well as the inevitable presence of added nuisances such as camera blur and noise in captured images.

Broadly, image upsampling methods can be divided into  {\em parametric} and {\em non-parametric} approaches.
Methods belonging to the first category assume parametric models for the unknown high-resolution image. The prominent method in this class is the familiar bicubic interpolation methods (that assumes a bandlimited structure on images), and is indeed the standard algorithm in commercial photo editing packages such as Adobe Photoshop.
Over the last 10 years, more sophisticated parametric models have been developed. Total variation minimization methods assume that images have bounded TV-norms~\cite{TV}); probabilistic models assume a prior on gradient profiles of images~\cite{nedi,gpp,fattal07}), and specify (or estimate from data) the hyperparameters of these priors. Sparse models assume that image patches are sparse in a basis or learnt dictionary~\cite{elad,yang}).

Methods belonging to the second category do not assume an explicit model for images or image patches; instead they exploit natural image invariances such as scale- or translation-invariance; in such cases, the prior for the high-resolution unknown image patches are raw patches from a database of external images~\cite{freeman}, or even patches from the scale-space input image~\cite{ebrahimi,irani}. These methods `learn' correspondences between high-resolution and low resolution patches and store this in a searchable database; during reconstruction, every occurrence of a low-resolution patch is replaced by its corresponding high-resolution patch.

Both types of approaches typically possess one or more of the following drawbacks: \\
{\bf Loss of fine details} Neither category of methods seems to reproduce realistic textural details for even moderate upscaling factors. \\
{\bf Visual artifacts} The resulting images suffer from various artifacts such as blurry edges, exaggerated noise, halos, ringing artifacts, and staircases (``jaggies"). \\
{\bf Computational demands} All state-of-the-art methods involve considerable computational costs. Upsampling a small ($128 \times 128$) image by a factor of 4 might typically take several minutes.

In this work, we develop a new hybrid algorithm for image upsampling which greatly alleviates each of these concerns. The key idea behind our approach is that {\em sharp edges and fine details in natural images exhibit inherently different structure}, and hence should be separately upscaled. We model the output image as the linear sum of edge and detail layers and pose the upsamping problem as a highly underdetermined linear inverse system of equations. We regularize this problem by imposing the image-specific prior that all patches of the target image can be well-approximated by the patches extracted from the low-resolution observed image~\cite{ebrahimi}. Thus, we obtain an optimization formulation of the upsampling framework.

We show that this optimization is highly non-convex, and indeed intractable.  To resolve this, we devise the following approximate algorithm. First, we use bicubic interpolation to obtain an initial high-resolution estimate. Next, we greedily replace each patch in the high-resolution image by a weighted sum of ``similar'' patches from the low-resolution patch manifold; we argue that this step both reconstructs the edge layer, as well as provides an $\alpha$-map indicating what pixels constitute the detail layer. Finally, to restore the fine textural details in the image, we decompose this intermediate image into its edge and detail layers using an edge-aware filtering technique~\cite{farbman}, and selectively enhance the detail layer using a parametric pixel-wise remapping function. See Fig.\ \ref{fig:analysis} for an illustration of our algorithm.

We demonstrate the utility of our method on a number of challenging standard test photos.
Our method consistently improves upon state-of-the-art results at much lower computational costs. The algorithm involves no training or learning of model parameters, and does not need an external database as input. Additionally, our method is fast and parallelizable, and lends itself to easy potential implementation for very high-definition images and videos.

\section{Layer-Based Upsampling}
\label{sec:model}

\subsection{Setup}

Consider a given low-resolution $n \times n$ image $I_0$. Our goal is to obtain a photorealistic high-resolution natural image $I$ whose downsampled version is equal to (or close to) $I_0$. The image $I$ can be conceptually decomposed into edge and detail layers:
\begin{equation}
I = E + D ,
\end{equation}
where the edge layer $E$ is piecewise smooth, and the detail layer $D$ consists of high-frequency variations of small magnitude. Figure~\ref{fig:sed} provides a simplified illustration of this model in 1D. If $\cal{L}$ denotes the linear downsampling operator, then the observed low-resolution image $I$ is given by:
\begin{equation}
I_0 = \mathcal{L}(I) = \mathcal{L} (E + D) .
\label{eq:samp}
\end{equation}
Equation \ref{eq:samp} represents a highly underdetermined system of linear equations comprising $n^2$ constraints and $2 \alpha^2 n^2$ unknowns, where $\alpha > 1$ is the desired upsampling factor. Consequently, this formulation is ill-posed.

\subsection{Optimization via the Patch Transform}

How do we regularize this problem? Given an image $I$, denote the {\em patch transform} ${P}(I)$ as the set of (overlapping) $r \times r$ patches extracted from the image.  Obviously, the patch transform ${P}$ is an invertible mapping; ignoring image boundary effects, every pixel in the image occurs in $r^2$ elements in $P$.
For completeness, we denote the (inverse) linear synthesis as $I = \mathcal{S} (P)$.
We trivially observe that any pixel intensity at a location $x$ in the image $I$ can be represented as the {\em mean} of the intensity values of the $r^2$ patches that contain the pixel ${\bf x}$, i.e., $I({\bf x}) = \textrm{mean}(p({\bf x})),~~p \in {P}(I)$. We also trivially observe that the {\em variance} of the $r^2$ different explanations of the pixel intensity (in short, the pixel variance) at $\mathbf{x}$ is zero.

Due to the one-to-one nature of the patch transform, any prior on the patch transform of an image is equivalent to a modified prior on the image itself. Then, an intuitive regularization for (~\ref{eq:samp}) would be to require that the patch transform ${P}(I)$ should be a subset of the {\em set of natural image patches} $\cal{M}$, since it is well known that the set $\cal{M}$ occupies an extremely tiny fraction of the space of all possible $r \times r$ patches. However, this constraint itself is unwieldy since there exists no known concise parametric or non-parametric model for $\cal{M}$; indeed there is no consensus for what characterizes a natural image.

Therefore, we adopt a more tractable relaxation of (\ref{eq:samp}): the patch transform of the target image ${P}(I)$ {\em must be contained in the patch transform of the low-resolution image} ${P}(I_0)$.
This hypothesis is justified by several recent important findings in the literature~\cite{irani11,fattal11} that natural image patches tend to be self-similar in highly localized regions in scale space. Therefore, our desired upsampled image can be posed as the solution to the following optimization:
\begin{eqnarray}
\widehat{I}&=&\arg \min \| I_0 - \mathcal{L}(I) \|, \\
 & &\textrm{s.t.}~~I = \mathcal{S}(P),~~\textrm{and}~~{P}(I) \subset {P}(I_0) \nonumber.
\label{eq:opt}
\end{eqnarray}

\subsection{Challenges}
The above optimization poses two primary caveats. The first caveat is computation complexity. Equation (\ref{eq:opt}) is a highly non-convex, combinatorial subset selection problem, and therefore is very difficult to optimize exactly. However, there potentially exist fast approximate schemes to obtain a suboptimal solution, one of which we will propose in Section~\ref{sec:method}.

The second caveat arises due to a subtle modeling error --- importantly in practice, that the patch transform of the low-resolution image $I_0$ only contains the patch transform of the edge layer $P(E)$, and cannot explain patches from the detail layer $D$. Indeed, as extensive experiments with natural images in~\cite{irani11} have shown, the scale invariance property for natural images holds for smooth regions and sharp edges, but not for fine textural details. When the detail layer $D = 0$, we can conceptually obtain a precise, consistent solution to (\ref{eq:opt}). However, when $D \neq 0$,  different patches in $P(\widehat{I})$ containing the location ${\bf x}$ would offer inconsistent explanations of the pixel value at ${\bf x}$. Consequently, the variance of the different pixel explanations at ${\bf x}$ is high, and the detail pixels gets ``averaged out''; this phenomenon is corroborated in our numerical experiments.

We propose a greedy nonparametric scheme to solve the above optimization problem. In the course of this scheme, we also detect which pixels possess high pixel variance; these correspond to the detail pixels. We then employ a simple detail enhancement heuristic to reconstruct the detail pixels, simultaneously making sure that the final result is consistent with the input low-resolution image. The full algorithm is as discussed below.

\begin{figure}[t]
\centering
\includegraphics[height=1in]{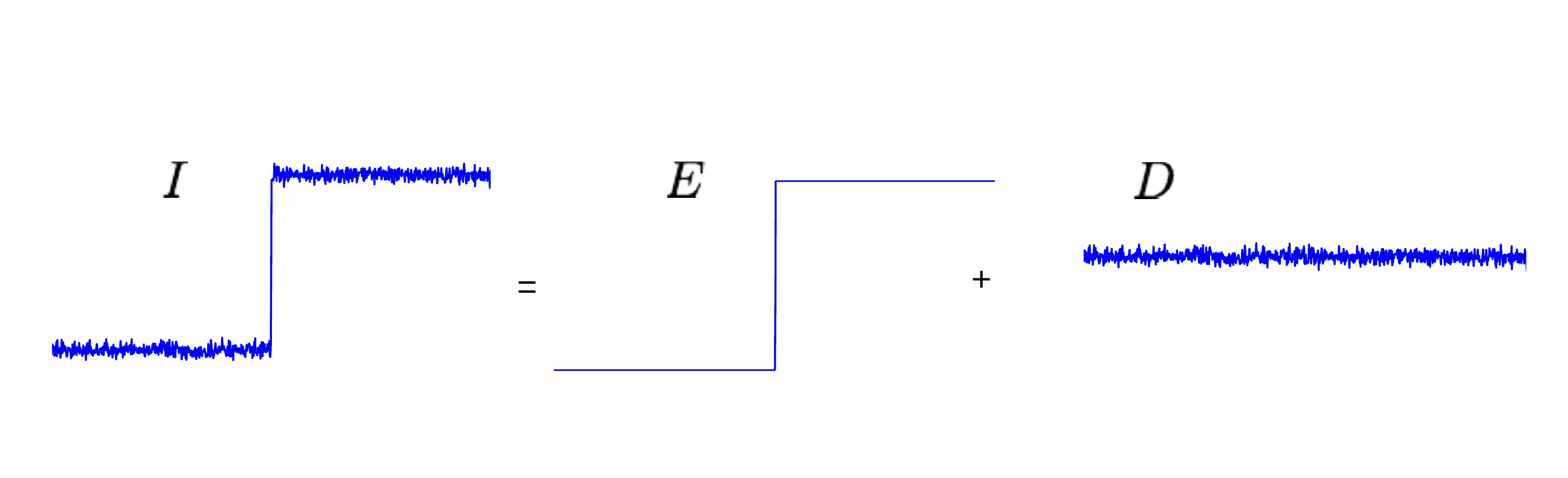}
\caption{Proposed image model illustrated in 1D. The image $I$ is modeled as the sum of an edge layer $E$ and a detail layer $D$.}
\label{fig:sed}
\end{figure}

\section{Proposed Upsampling Algorithm}
\label{sec:method}

\subsection{Description}

Our algorithm consists of two duelling subroutines that are applied in an alternating fashion. We first estimate the patch transform of the desired image using the non-parametric approach: given a blurry high-resolution patch, we extract a few similar raw patches from the input low-resolution image and replace the blurry patch with a weighted combination of the extracted raw patches.
Next, we perform an additional edge-detail decomposition of the estimated image, and carry out a pointwise remapping to enhance only the detail pixels. A careful recombination of the outputs of this decomposition results in a photorealistic final image that consists of sharp edges, fine scale details and no visible artifacts in smooth regions. For the rest of the paper, we assume that we are working with intensity (luminance) images with dynamic range $[0,1]$. \\
 \\
{\bf Estimation of the edge layer}
Given a low-resolution image $I_0$, we perform an initial upsampling by a small factor using a fast parametric  interpolation (we use bicubic interpolation in all our experiments.) If $\cal{U}$ denotes a linear interpolation operator, then the image
$$\check{I} = \mathcal{U}(I_0)$$
represents our initial intermediate estimate of the upsampled image. However, this parametric interpolation results in reduced sharpness of edges, resulting in a degradation in quality. We compensate for this as follows. Given a blurry high-resolution $r \times r$ patch $p({\bf x}) \subset \check{I}$ (where $\bf x$ denotes the spatial location), we perform a {\em localized search} for raw $r \times r$ patches in the given low-resolution image $I_0$ ($r = 5$ in all our experiments). Formally, we perform a nearest-neighbor search:
\begin{equation}
\label{eq:nn}
q = \arg \min_{p'({\bf x}') \in I_0} d_1(p',p) + \lambda d_2({\bf x}',{\bf x}),
\end{equation}
where $d_1$ is the distance between the patch intensity values and $d_2$ represents the distance between the spatial locations of the patches (normalized with respect to the size of the image) . We choose this distance function since it penalizes both In the standard manner, we remove the DC component of each patch before performing the nearest neighbor optimization. In our experiments, we set a value of $\lambda = 10$.

\begin{figure}[t]
\centering
\begin{tabular}{ccc}
\includegraphics[height=1in]{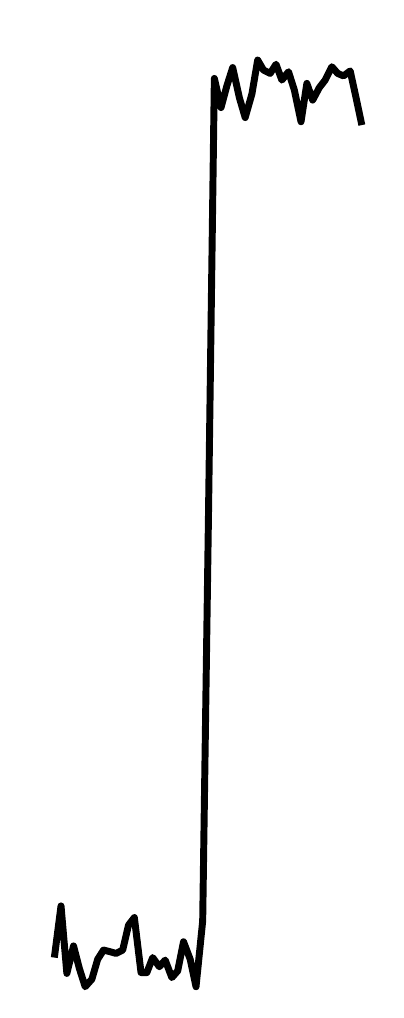} &
\includegraphics[height=1in]{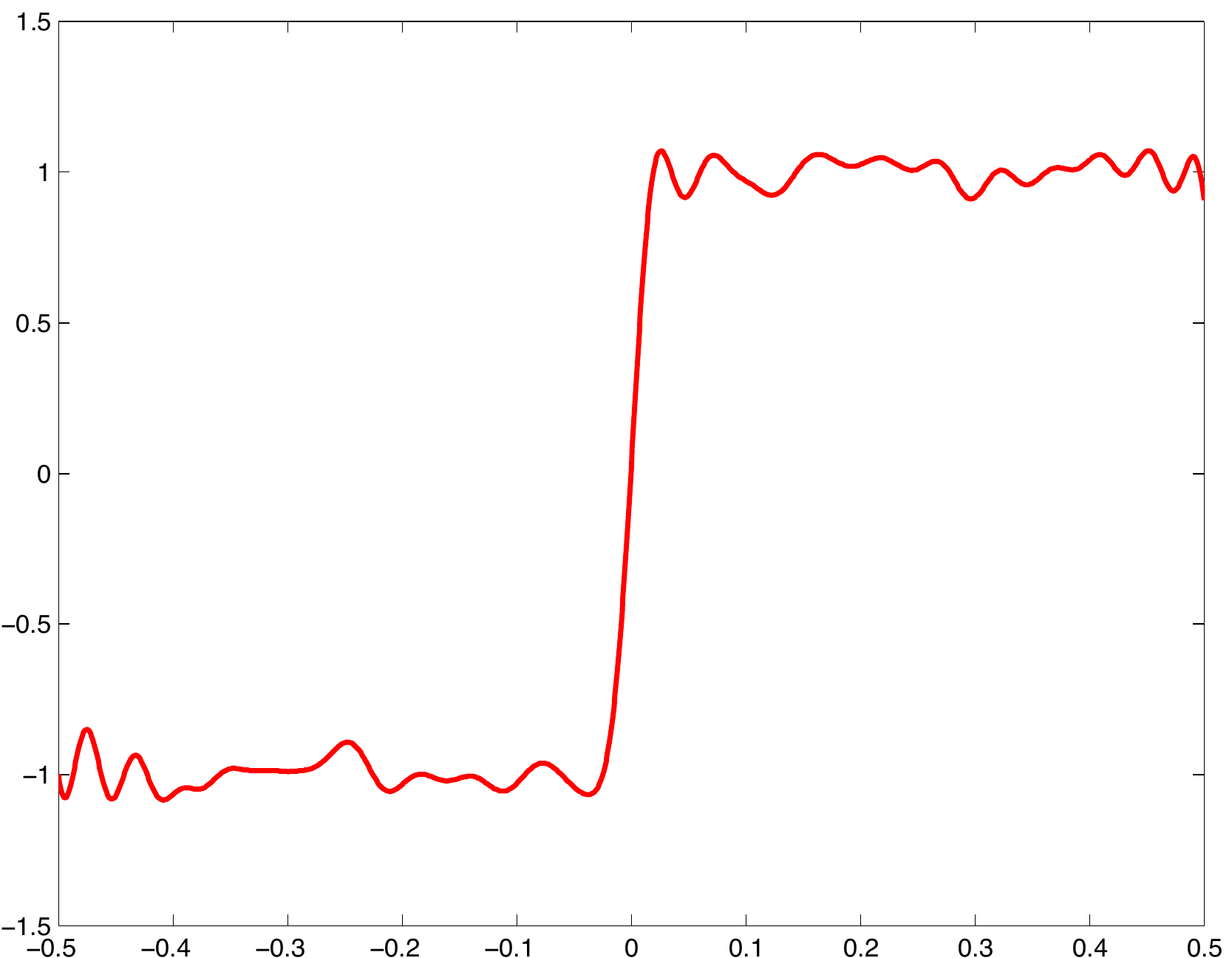} &
\includegraphics[height=1in]{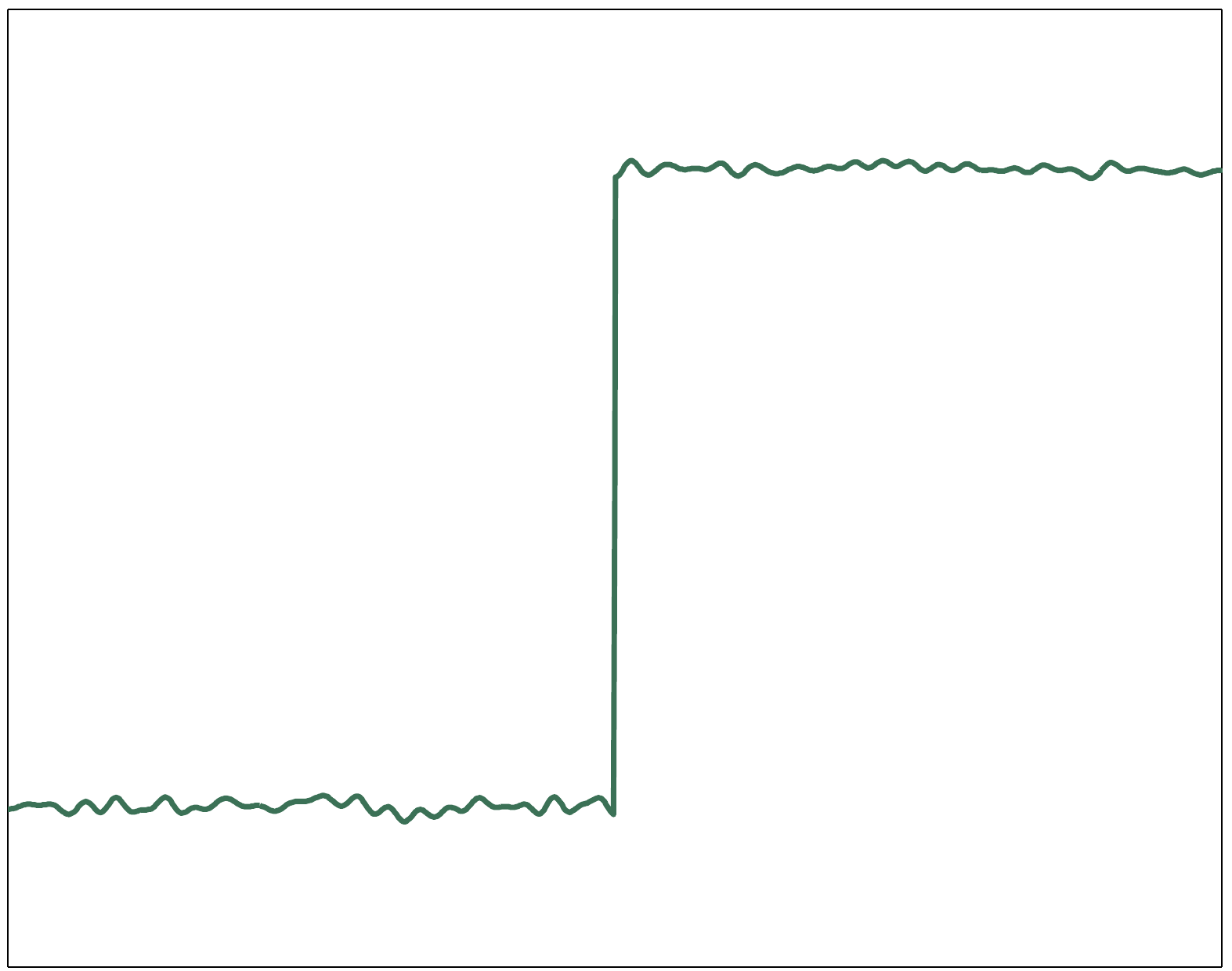} \\
$I_0$ & $\check{I}$ & $\tilde{I}$.
\end{tabular}
\caption{Various stages of the proposed upsampling process illustrated in 1D. The input image $I_0$, the downsampled version of the image $I$ in Fig.\ \ref{fig:sed}, consists of both sharp edges and detail patches. An initial bicubic interpolation provides an estimate $\check{I}$ which consists of blurry (unsharp) edges and reduced details. Replacing every patch in the patch transform of $\check{I}$ by its nearest neighbor in the patch transform of $I_0$ produces the nonparametric estimate $\tilde{I}$, which has sharp edges and reduced details.}
\label{fig:analysis1d}
\end{figure}

Once a few `close' example patches $q_j$ have been found,  we replace the patch $p$ with a {\em weighted} combination of the extracted raw patches. Overlaps between resulting patches $q$ are further averaged together to reduce blocking artifacts:
\begin{equation}
\label{eq:rec}
\tilde{I}({\bf x}') = \sum_{j : {\bf x}' \in q_j} q_j \cdot w_{q_j} \cdot G_\sigma,
\end{equation}
where the Gaussian weights $G_\sigma$ favor greater weights to the center of the $5 \times 5$ patches $q$ ($\sigma = 1.25$ in all our experiments), and the weights $w_{q} = d_1(q,p)^{-1}$ favor greater weights to better matches to the image. Note that Eq.\ \ref{eq:rec} effectively calculates a weighted version of the synthesis step $\cal{S}$, the inverse of the patch transform of $\tilde{I}$. Thus, if $\mathcal{N}$ denotes the overall non-parametric synthesis procedure, then we have that $$\tilde{I} = \mathcal{N}(\check{I}).$$ Thus, we obtain a second intermediate image $\tilde{I}$ with sharp, realistic-looking edges.

\begin{figure}[t]
\centering
\includegraphics[width=0.6\linewidth]{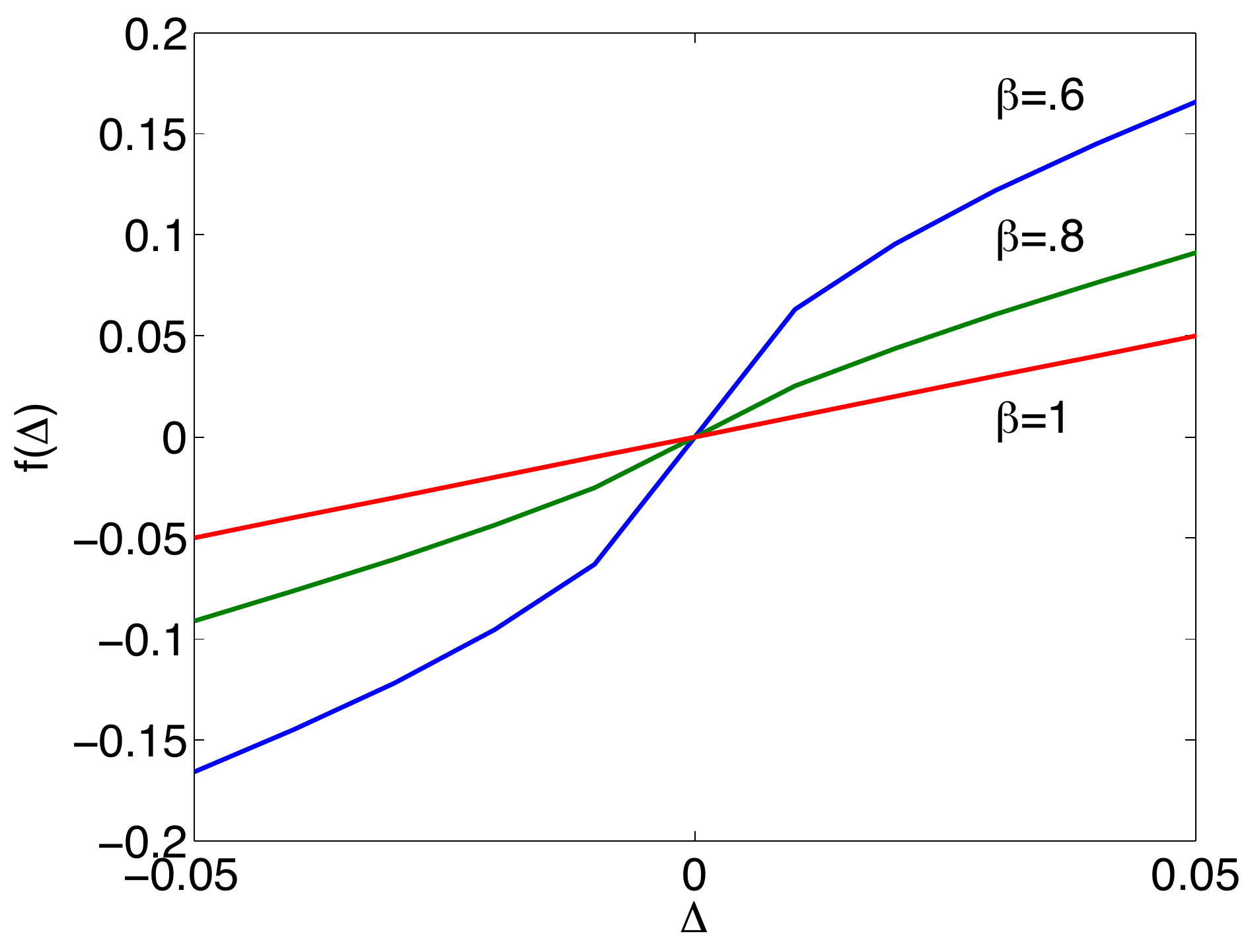}
\caption{Detail-enhancing pointwise pixel remapping functions $f(\Delta) = \Delta^\beta$ for various values of $\beta$. Smaller values of $\beta$ indicate a higher amount of detail enhancement.}
\label{fig:scurve}
\end{figure}

In this manner, we have {\em simulated} the optimization in Eq.\ \ref{eq:opt}. However, as argued, the obtained estimate $\tilde{I}$ suffers from a loss of detail in textured regions.  The reason is that the small, fine-scale variations present in textured regions in the high-resolution target image do not manifest themselves in their lower-resolution counterparts. Therefore, they cannot be synthesized by a simple non-parametric search and are averaged out in the synthesis step $\cal{S}$ (Eq.~\ref{eq:rec}). \\
\\
{\bf Estimation of the detail layer}
To solve this problem, we perform a global detail enhancement on $\tilde{I}$. We decompose $\tilde{I}$ into a piecewise-smooth component and a detail component, akin to the model shown in Fig.\ \ref{fig:sed}, using the weighted least-squares (WLS) approach~\cite{farbman}\footnote{We chose this method since it generates high quality decompositions in linear time. Other decompositions, such as the bilateral filter~\cite{bilateral} may equally be used.}: $\tilde{I} = \tilde{E} + \tilde{D}$. Next, we apply a detail-enhancing pixelwise remapping function:
$$ f ( \Delta ) = \Delta^{\beta},$$
to obtain a new detail layer image $f( \tilde{D} )$. Informally, we call this function an ``S-curve''; see Fig.\ \ref{fig:scurve} for some example remapping functions $f(\cdot)$.   We will discuss how to choose the scalar parameter $\beta$ below.

This step restores fine-scale details, but also aggravates noise and artifacts in smooth and edge regions. Therefore, while synthesizing the final image, we only consider the pixels that have been ``excessively averaged'' in the previous step. This is easily achieved by computing the variance of different explanations for a pixel in the patch transform synthesis (Eq.~\ref{eq:rec}), and constructing a binary $\alpha$-mask that indicates the pixels with high pixel variance (Fig. \ref{fig:analysis}).  We selectively blend in these pixels from the enhanced detail layer using the equation:
$$
\widehat{I} = (1 - \alpha) \tilde{I} + \alpha f (\tilde{D}) = \mathcal{E}(\tilde{I}),
$$
where $\mathcal{E}$ denotes the overall detail-enhancement operator. This final step results in a high-resolution image $\widehat{I}$ with sharp edges and a requisite amount of details.  Thus, the overall upscaling procedure can be summarized by the composite operator relation:
\begin{equation}
\label{eq:comp}
\widehat{I} = \mathcal{E} ( \mathcal{N} ( \mathcal{U} (I_0))) .
\end{equation}

\begin{figure*}[th]
\begin{tabular}{ccccc}
\centering
   \includegraphics[trim = 25mm 40mm 25mm 60mm, clip, width=1.2in]{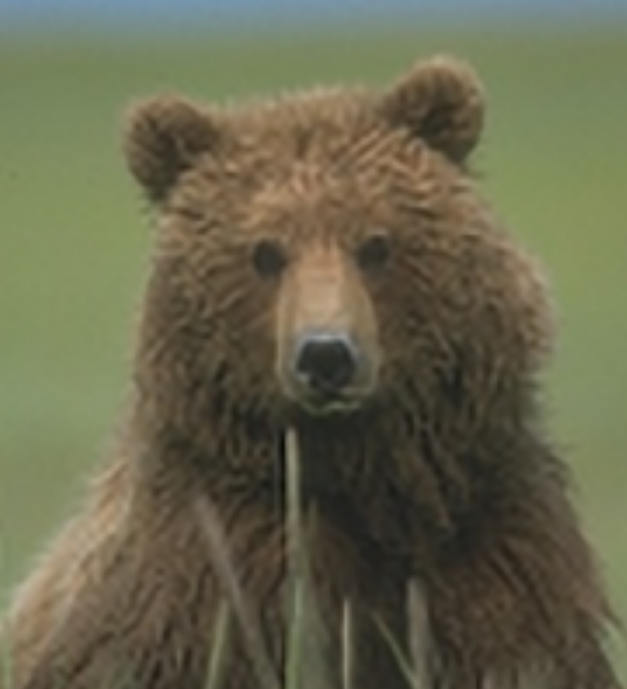} &
   \includegraphics[trim = 25mm 40mm 25mm 60mm, clip, width=1.2in]{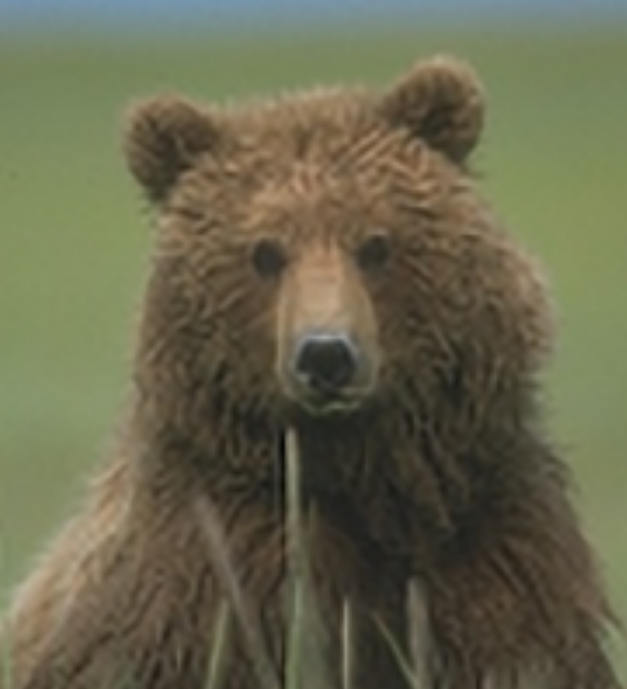} &
   \includegraphics[trim = 25mm 40mm 25mm 60mm, clip, width=1.2in]{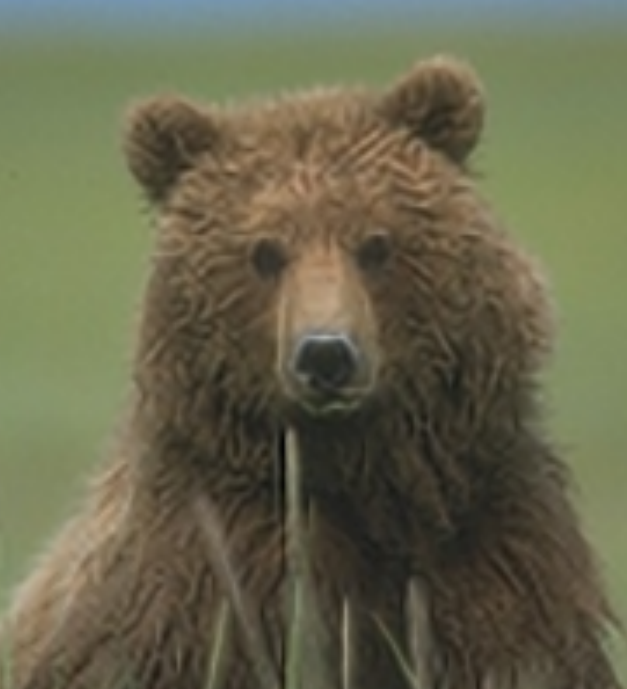} &
   \includegraphics[trim = 25mm 40mm 25mm 60mm, clip, width=1.2in]{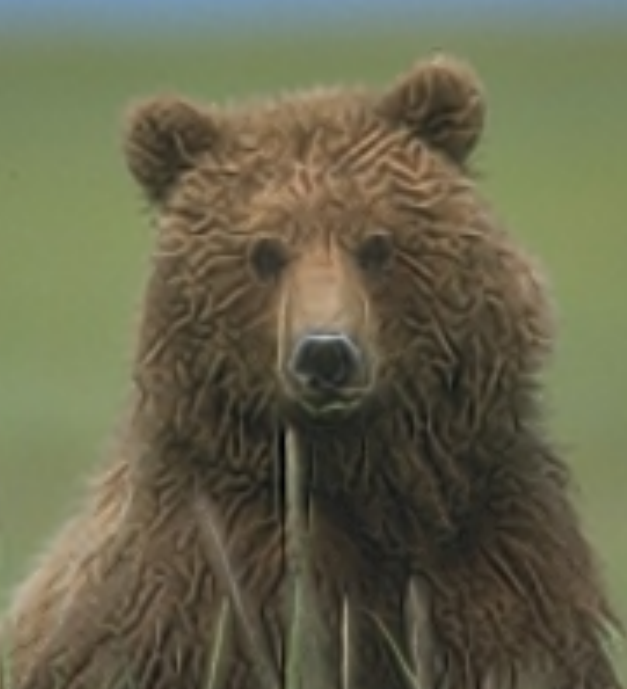} &
   \includegraphics[trim = 25mm 40mm 25mm 60mm, clip, width=1.2in]{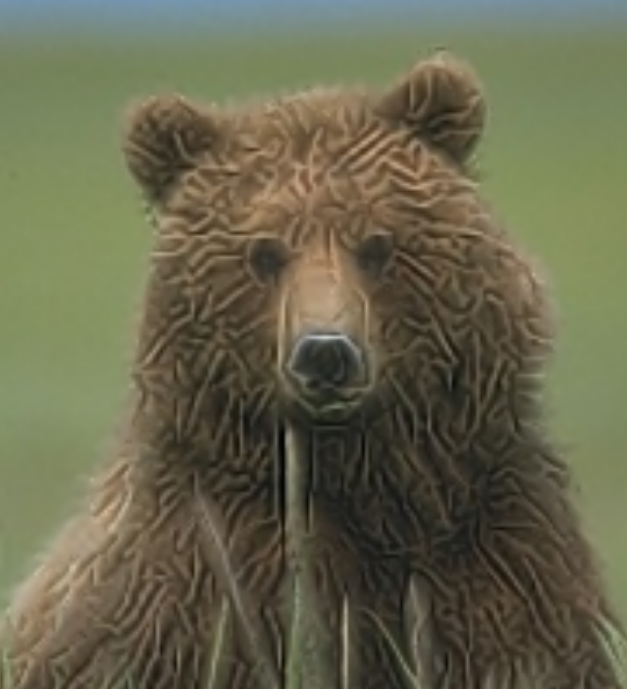} \\
{\small $\beta = 1.0$} & {\small $\beta = 0.9$} & {\small $\beta = 0.8$} & {\small $\beta = 0.7$} & {\small $\beta = 0.6$}
\end{tabular}
   \caption{Example (cropped) $3\times$ upsampling results of a $240\times160$ image using layer-based upsampling. A value of $\beta = 0.8$ gives visually pleasant results for a large set of images.}
\label{fig:bear}
\end{figure*}


Figure~\ref{fig:analysis1d} illustrates (on a simplified 1D signal) the results of some intermediate steps in our proposed algorithm. The bicubic upsampling step $\mathcal{U}$ yields poor results on both edges and textures; the non-parametric synthesis step $\mathcal{N}$ corrects for the edges, but suffers from a lack of details.
Figure~\ref{fig:analysis} illustrates the results on a standard test image, and serves to highlight the importanc of the detail enhancement step $\mathcal{E}$ within our framework. The step $\mathcal{E}$, together with an intuitively chosen blending function $\alpha$, ensures that relevant and photorealistic details are added.\\
\\
{\bf Consistency}
An important consideration is to ensure that each of the upsampling operators results in an image that is consistent with the low-resolution observed image. We achieve this by running a few iterations of the linear back-projection step~\cite{backpr} (commonly used in many upsampling schemes) for each intermediate estimate $\tilde{I}$ and $\widehat{I}$. This step is equivalent to finding the closest image in the subspace of feasible high resolution solutions to the intermediate image, thus ensuring a final estimate $\widehat{I}$ that is consistent with the low resolution input image $I_0$.

\subsection{Discussion}
\label{subsec:discuss}

{\bf Implementation details}
In practice, the above upsampling process gives desirable results for upscaling factors $< 1.25$. This is due to the fact that the scale-invariance hypothesis is truly valid only for small scale factors.
To achieve any desired magnification factor, our procedure is repeated several times (e.g., to achieve a upscaling factor of 4, the above three-step approach is performed by applying a scale factor of $\gamma = 2^{1/3}$ a total of 6 successive times.)

In all our experiments, we perform the above upsampling by first transforming the image into YCbCr space, and applying the upsampling to the luminance channel. This is due to the well-known fact that visual perception of edges and details responds primarily to the luminance of the image. \\
\\
{\bf Parameters}
In sharp contrast with parametric upsampling methods such as imposed edge statistics~\cite{fattal07} and gradient profile priors~\cite{gpp}, our method does not involve any training phase involving estimation of different parameters. In our experiments, we have observed that our method is robust to the choice of the number of intermediate upsampling steps $\gamma$; further, it is also robust to the choice of the number of nearest neighbors obtained in the nonparametric search $k$, as well as the choice of the weighting parameter $\lambda$ that controls the tradeoff between accuracy of nearest neighbors versus the locality of the search. The only changing parameter in our different experiments is the exponent $\beta$ used in the detail enhancement step; this varies depending on the amount of detail present in the original image, as well as the sharpness of desired detail in the upsampled image. A smaller value of $\beta$ corresponds to a `more-acute' S-curve, and hence exaggerates finer details. Figure~\ref{fig:bear} illustrates the effect of varying the parameter $\beta$ across a range of values ($\beta = 1.0$ implies that no detail layer exaggeration is performed). For small values of $\beta$, we notice that the details get artificially exaggerated. Unless otherwise stated, we use the value $\beta =  0.8$ in our experiments; in our experience, this choice yields visually pleasant results.\\
\\
{\bf Computational advantages}
Our proposed method does not involve any external database; instead, it only relies on self similarities present in the input image. Further, it requires no special learning, training, or tuning of model parameters. Moreover, our method obviates the need for an exhaustive search of the image's scale space. To synthesize each upsampling estimate, we only operate upon pixels and patches belonging to the current intermediate estimate.
The dominant factor in computational complexity is the nearest neighbor search (Eq.~\ref{eq:nn}). A higher value of $\lambda$ in the minimization implies two axis aligned orthogonal principle search directions (two spatial coordinates) in the search space. Therefore exact nearest neighbor search for a given patch can be performed in $\mathcal{O}( N^{\frac{1}{2}})$ time using adaptive k-d tree search structure where $N$ is the number of pixels in the image. This yields sub-quadratic complexity $\mathcal{O}( N^{\frac{3}{2}})$ for the overall algorithm, which is a significant improvement over state-of-the-art nonparametric algorithms~\cite{irani}, which involve a complex training phase and whose best-case complexity is quadratic ($\mathcal{O}(N^2)$).

\begin{figure*}[t]
\begin{tabular}{ccc}
\centering
   \includegraphics[width=2.15in]{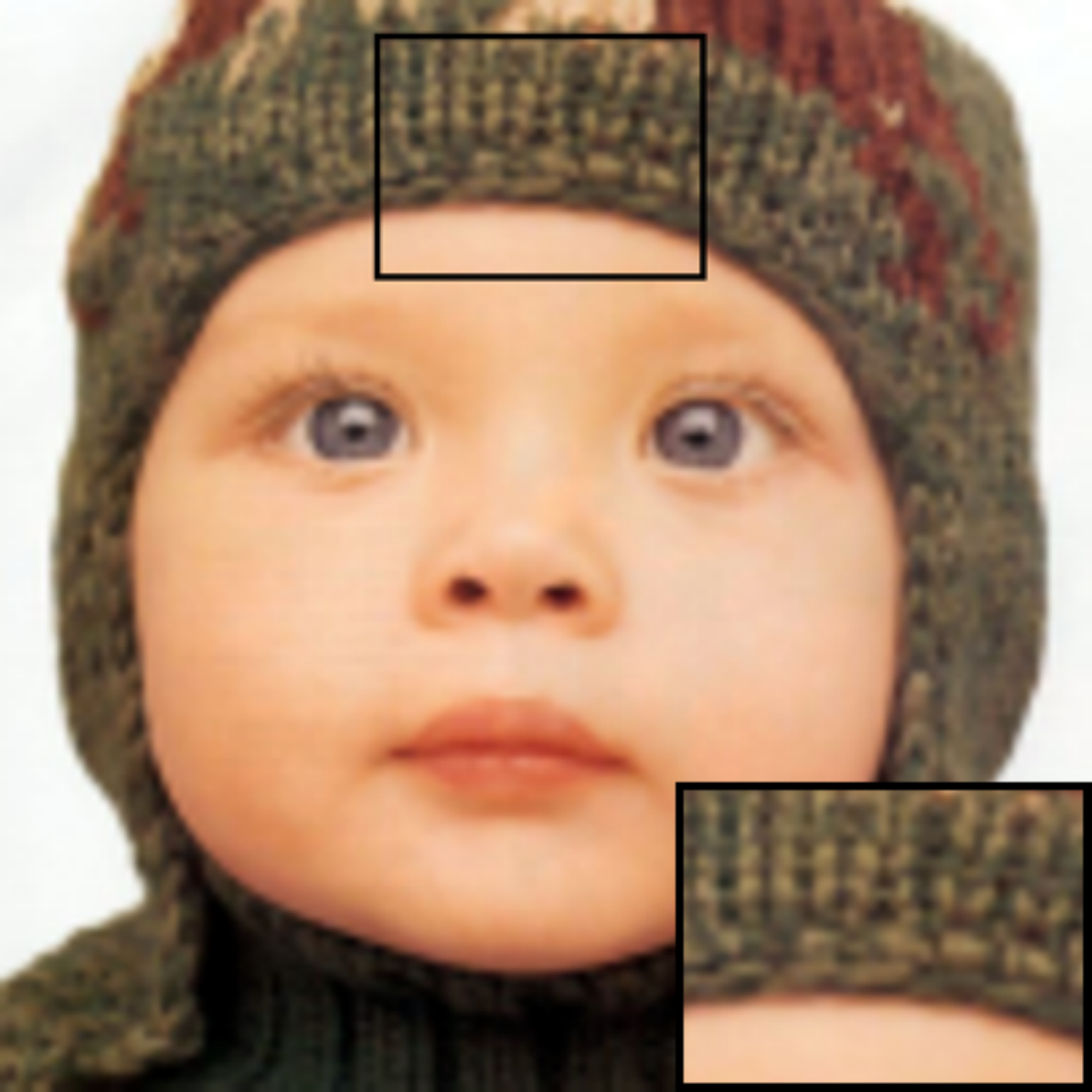} &
   \includegraphics[width=2.15in]{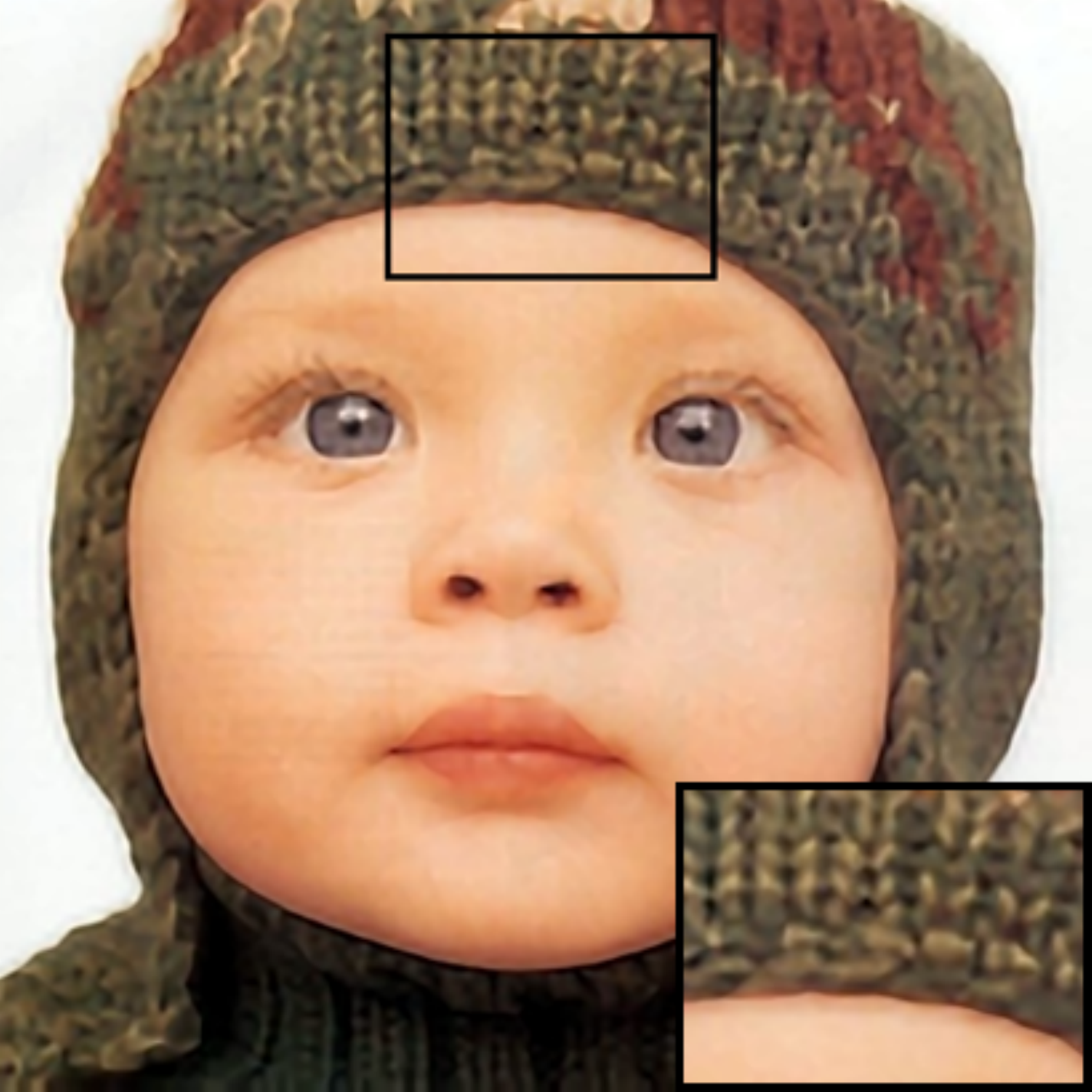} &
   \includegraphics[width=2.15in]{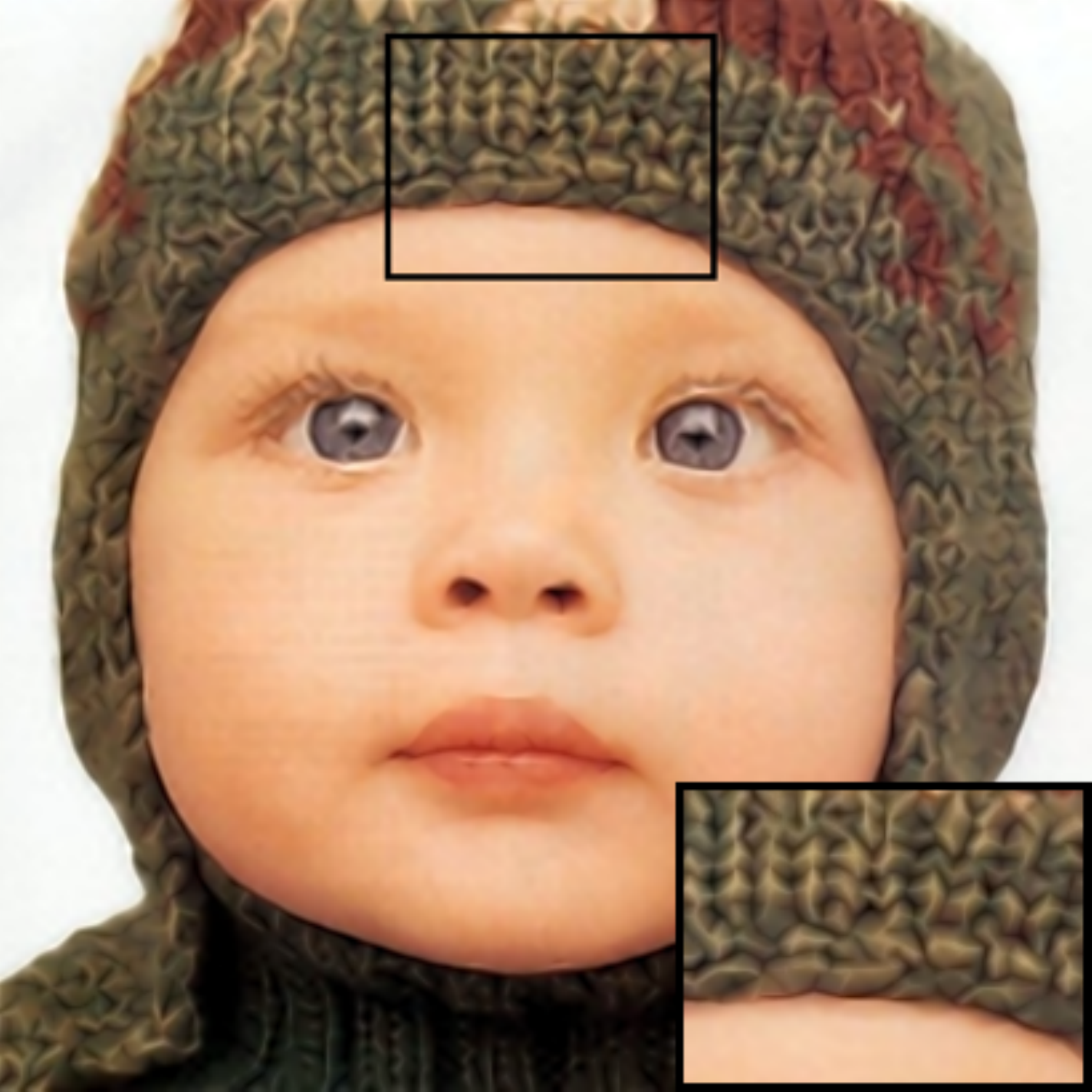} \\
{\small Bicubic interpolation} & {\small Scale-space search~\cite{irani}} & {\small Proposed layer-based method}
\end{tabular}
   \caption{Example $4\times$ upsampling result of a  $128\times128$ image using layer-based upsampling. Our proposed method retains sharpness of salient edges, eliminates halos, as well as reproduces fine textural details at greatly reduced computational cost. Quality of images in this paper best evident when viewed full-screen on a computer monitor.}
\label{fig:child1}
\end{figure*}

\begin{figure*}[!t]
\begin{tabular}{ccc}
\centering
   \includegraphics[trim = 65mm 0mm 0mm 40mm, clip, width=2.15in]{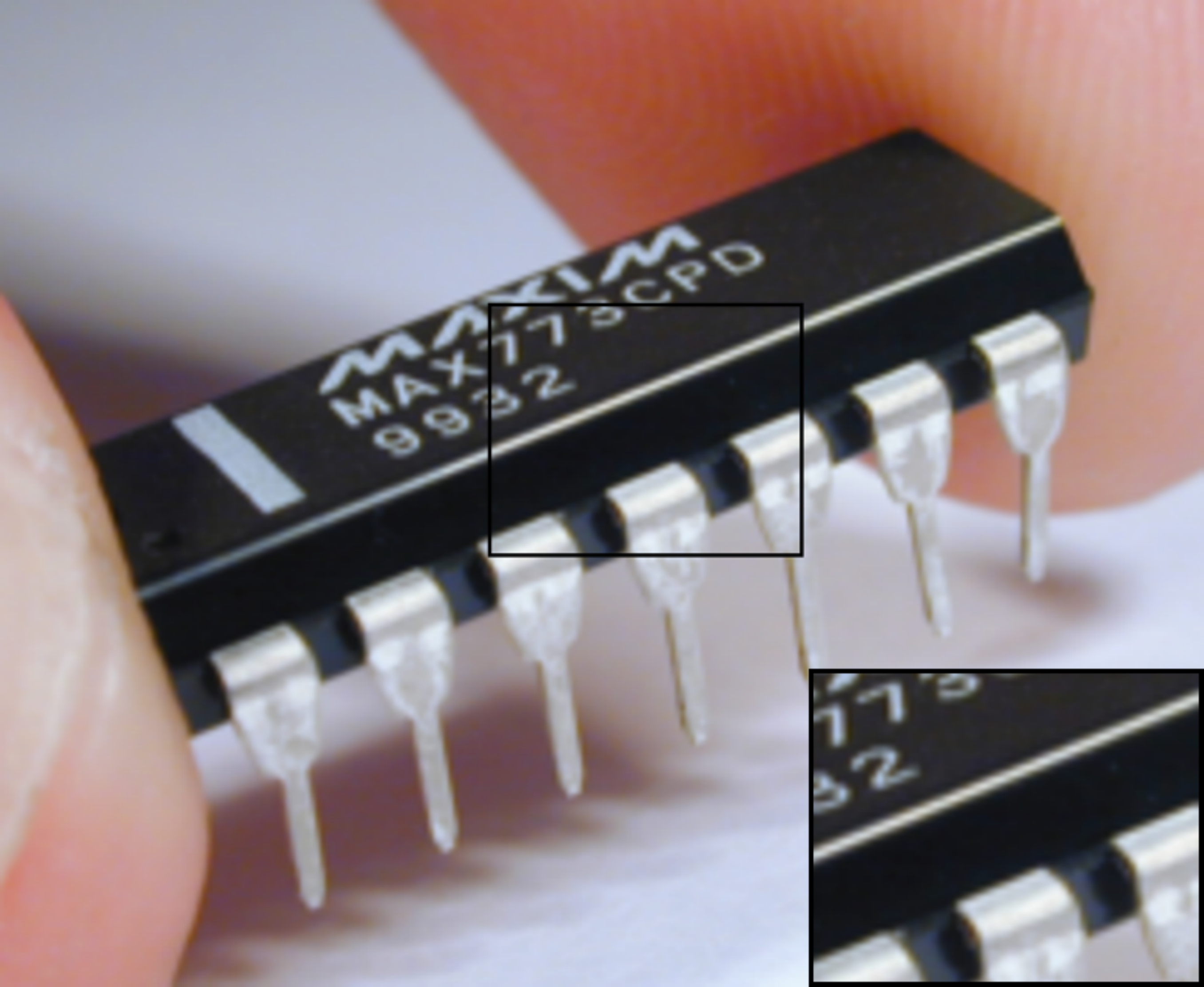} &
   \includegraphics[trim = 65mm 0mm 0mm 40mm, clip, width=2.15in]{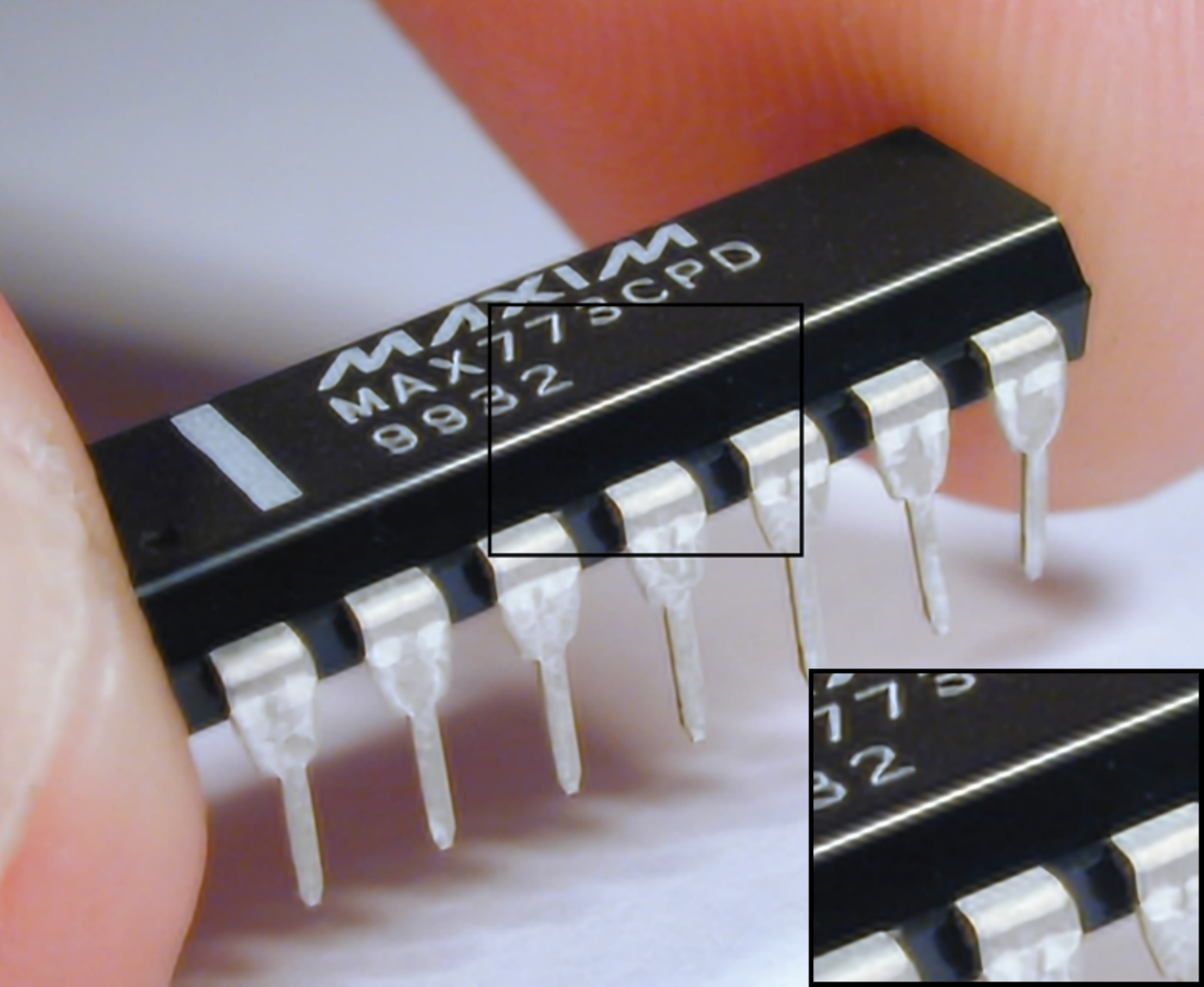} &
   \includegraphics[trim = 65mm 0mm 0mm 40mm, clip, width=2.15in]{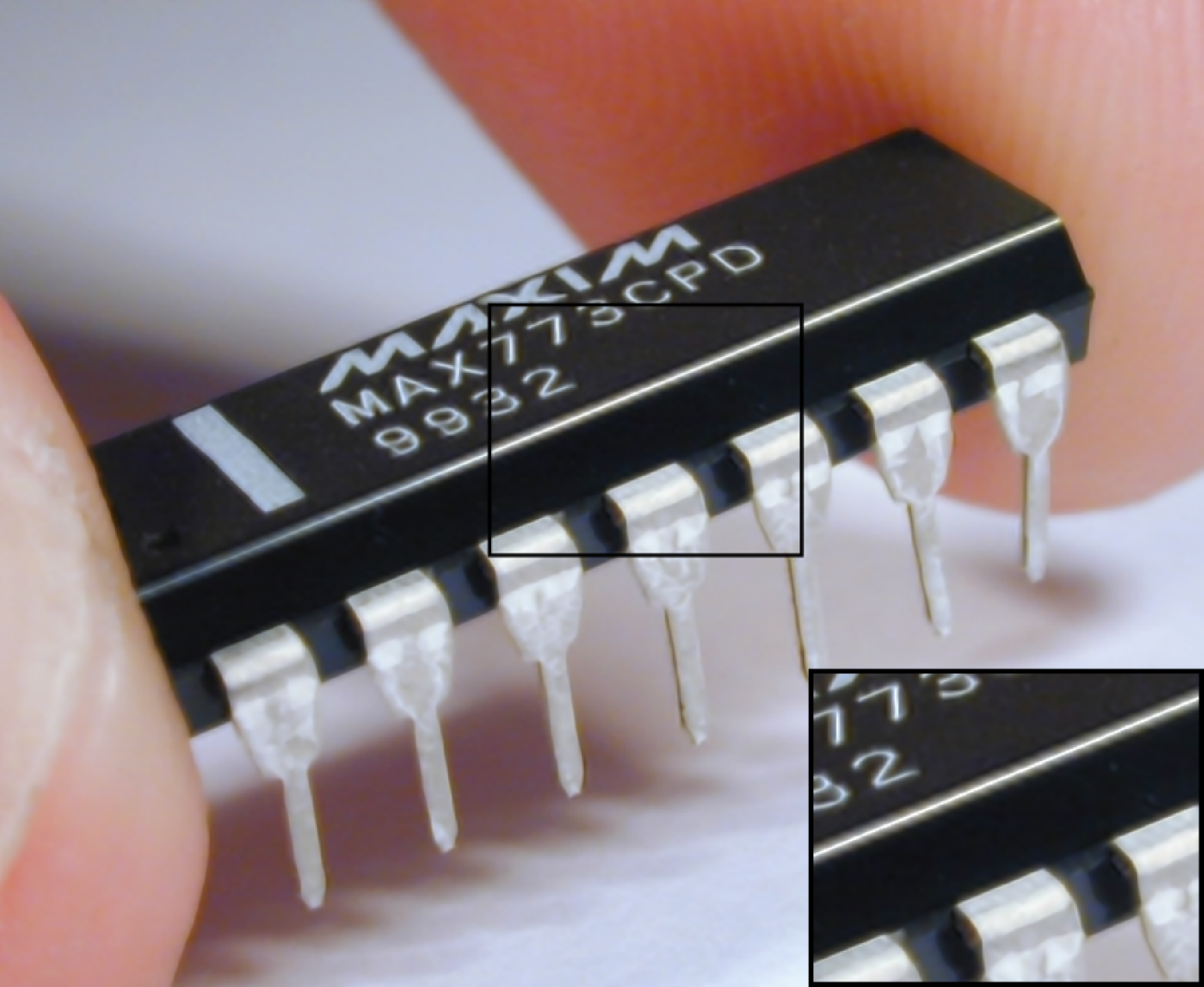} \\
{\small Bicubic interpolation} & {\small Scale-space search~\cite{irani}} & {\small Proposed layer-based method}
\end{tabular}
   \caption{Example $4\times$ (cropped) upsampling results of a $244\times200$ image using layer-based upsampling.}
\label{fig:chip}
\end{figure*}

\section{Results}
\label{sec:exp}
\begin{figure*}[t]
\begin{tabular}{ccc}
\centering
   \includegraphics[height=2.15in]{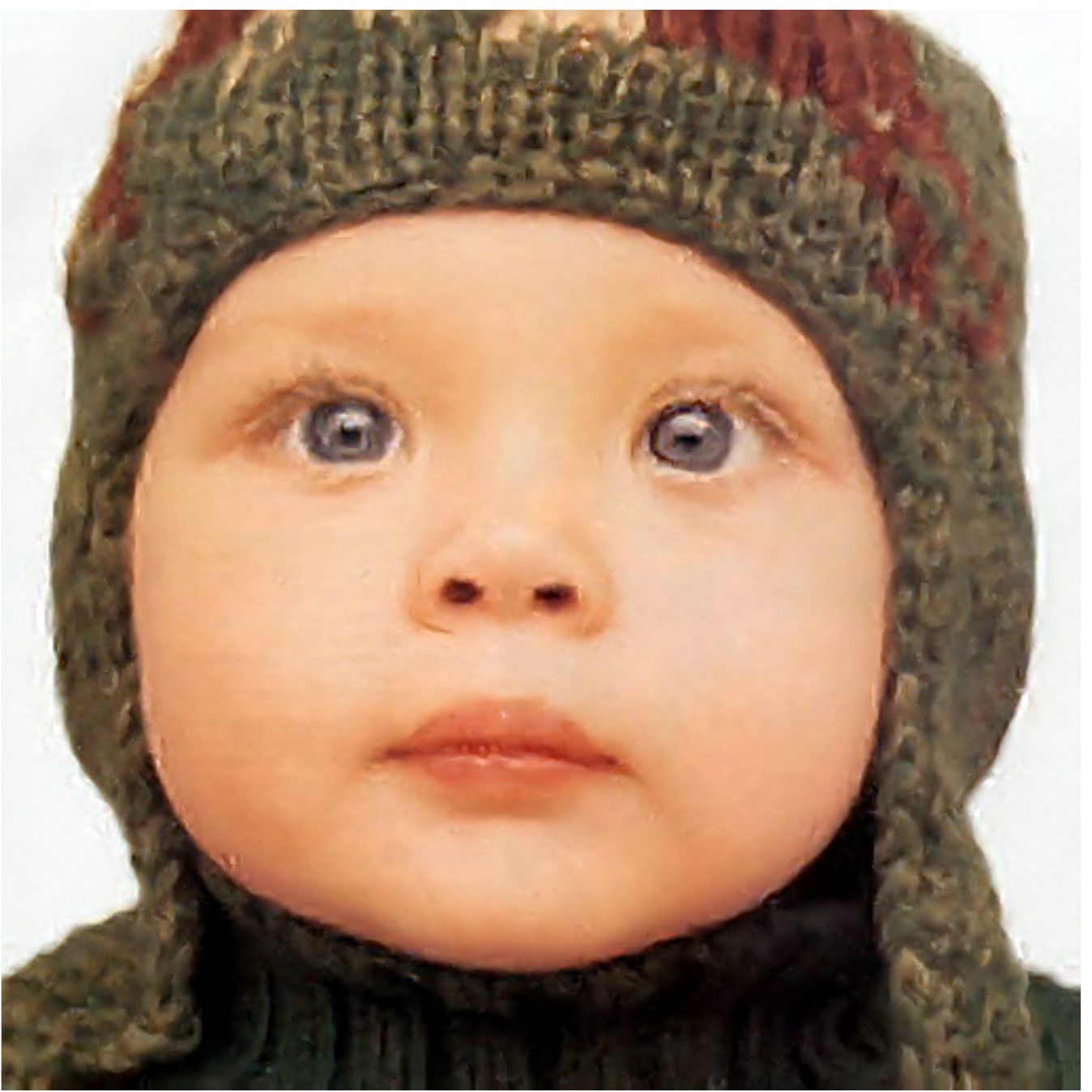} &
   \includegraphics[height=2.15in]{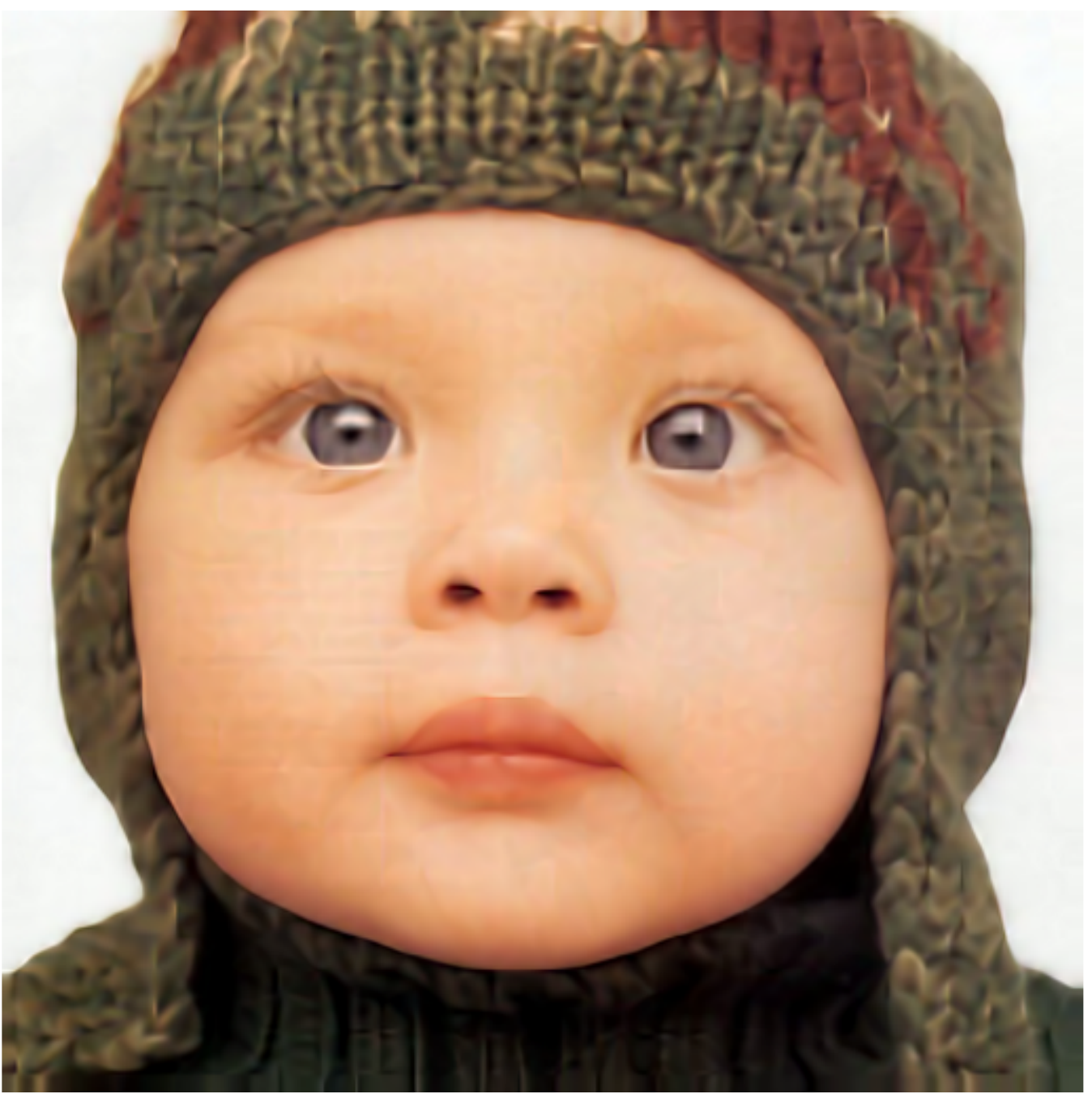} &
   \includegraphics[height=2.15in]{child_4x_ours2} \\
{\small Example-based SR~\cite{freeman}} & {\small Filter-based upsampling~\cite{fattal11}} & {\small Proposed layer-based method}
\end{tabular}
   \caption{Example $4\times$ upsampling results of a $128\times128$ image using layer-based upsampling: additional comparisons.}
\label{fig:child2}
\end{figure*}

\begin{figure}[!t]
\centering
\begin{tabular}{cc}
\centering
   \includegraphics[trim = 35mm 20mm 35mm 20mm, clip, width=1.5in]{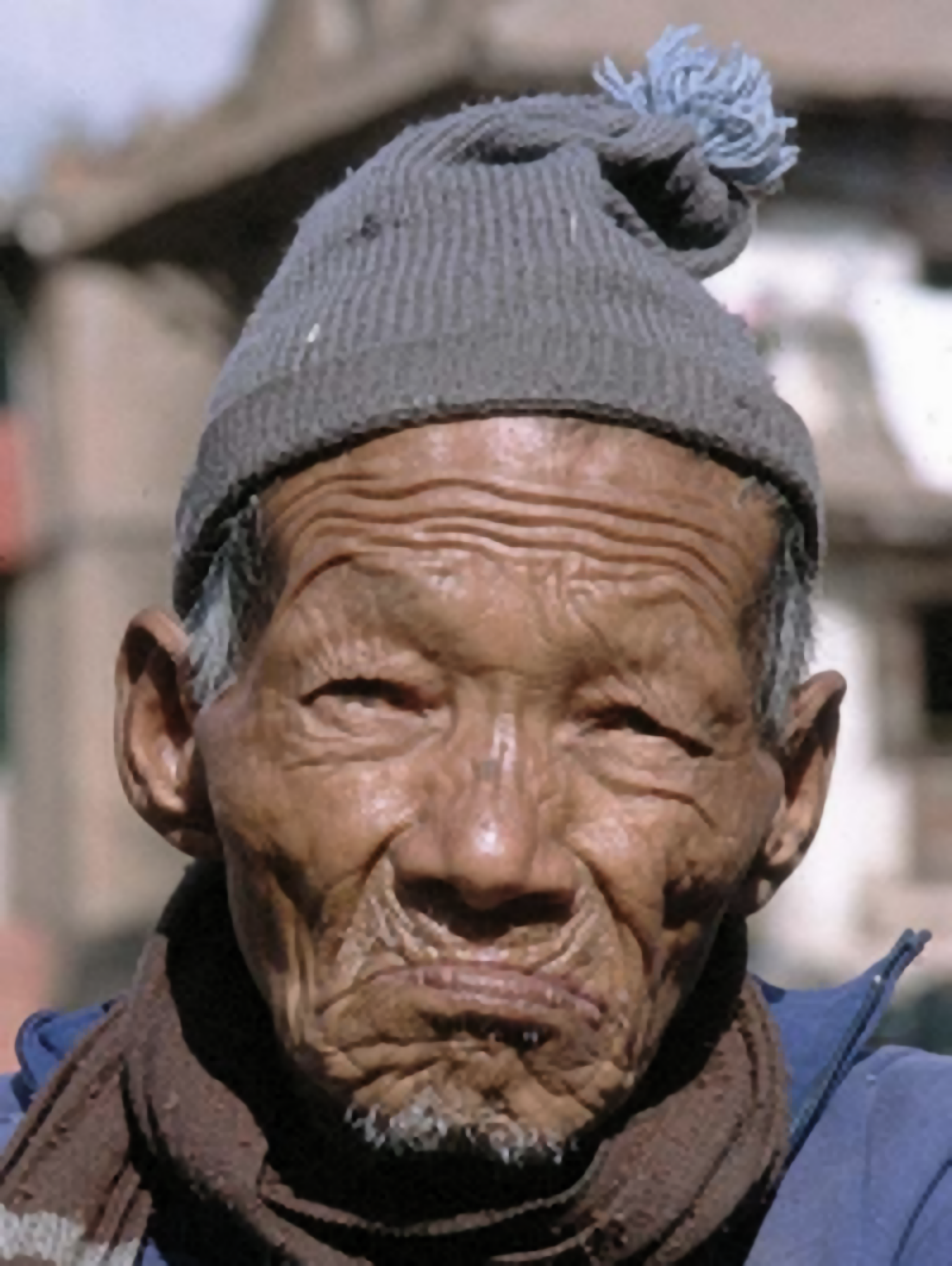} &
   \includegraphics[trim = 35mm 20mm 35mm 20mm, clip, width=1.5in]{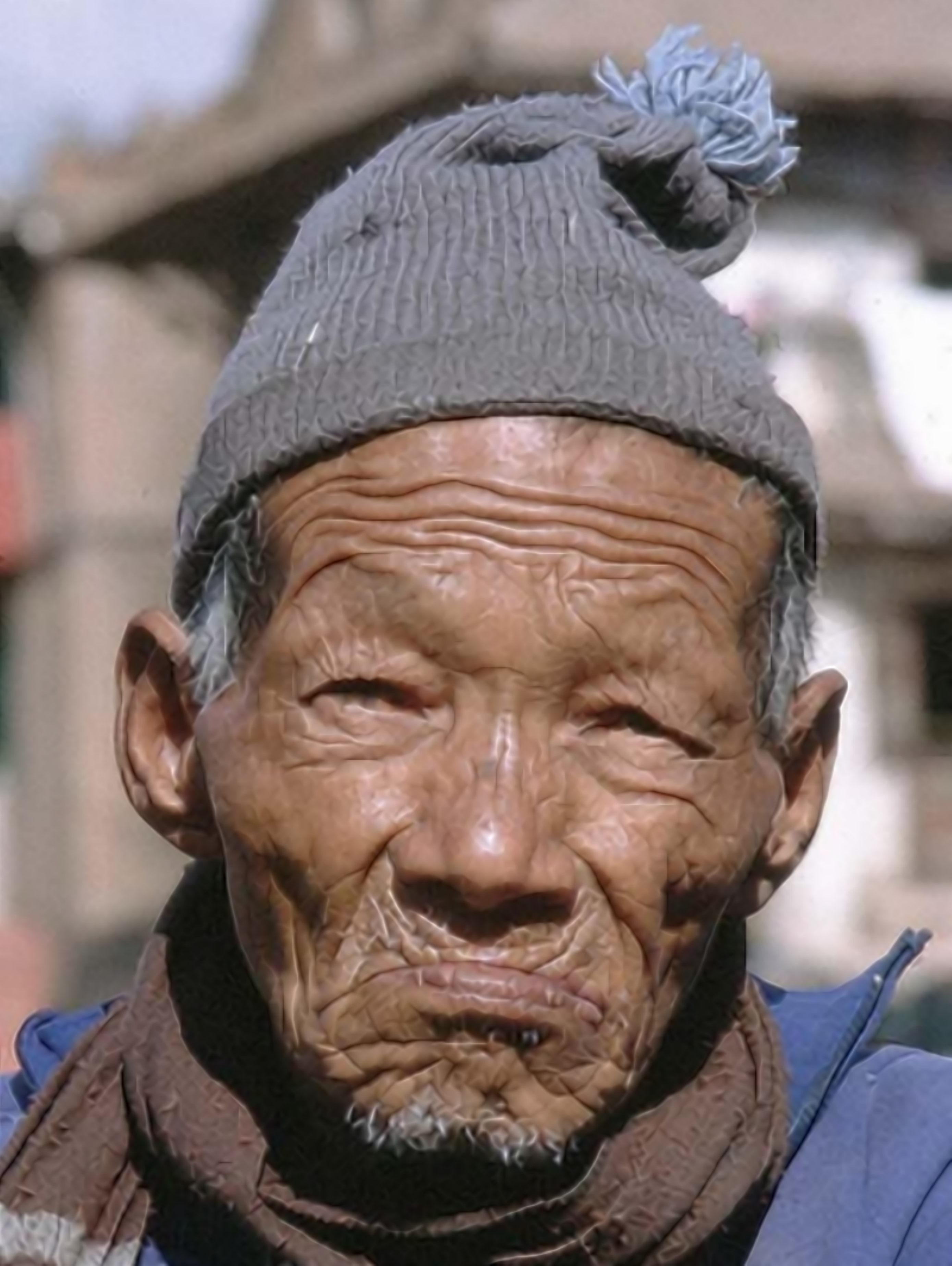} \\
{\small Imposed edge statistics~\cite{fattal07}} &
{\small Proposed layer-based method}
\end{tabular}
   \caption{Example $4\times$ (cropped) upsampling results of a $266\times200$ image using layer-based upsampling.}
\label{fig:oldman}
\end{figure}

We test our algorithm on a number of standard images, and compare it with state-of-the-art parametric as well as example-based image upsampling algorithms~\cite{irani,fattal11}.
Figure~\ref{fig:child1} indicates the significant improvements achieved by our method over the nonparametric scale-space search method advocated by Glasner and Irani~\cite{irani}; our method accurately reproduces sharp edges, reduces haloing artifacts, and synthesizes photorealistic textures.
Figure~\ref{fig:child2} compares the performance of our algorithm with other state-of-the-art algorithms, such as example-based super-resolution by Freeman et al.~\cite{freeman} and edge-aware filter-based upsampling by Freedman and Fattal~\cite{fattal11}. The perceptual improvement offered by our method is evident, particularly in detail (texture) regions.
Figure~\ref{fig:oldman} illustrates the improved performance of our proposed algorithm over the method of Imposed Edge Statistics~\cite{fattal07}, simultaneously obviating the need for a complex parameter training phase, as well as much improved speed of recovery.

\begin{table}
\centering
\begin{tabular}{| c | c | c |}
\hline
Method & PSNR (dB) & SSIM \\
\hline
Example-based SR~\cite{freeman} & 21.43 & 0.6438 \\
Filter-based upsampling~\cite{fattal11} & 21.30  &  0.6483 \\
Scale-space search~\cite{irani} & 21.45 & 0.6829 \\
Proposed layer-based method & {\bf 23.93} & {\bf 0.7624} \\
\hline
\end{tabular}
\vspace{.5mm}
\caption{Numerical comparison of upsampling algorithms for the images in Figs~\ref{fig:child1} and~\ref{fig:child2}.}
\label{tab:comparisons}
\end{table}

\begin{figure*}[!t]
\begin{tabular}{cccc}
\centering
   \includegraphics[width=1.5in]{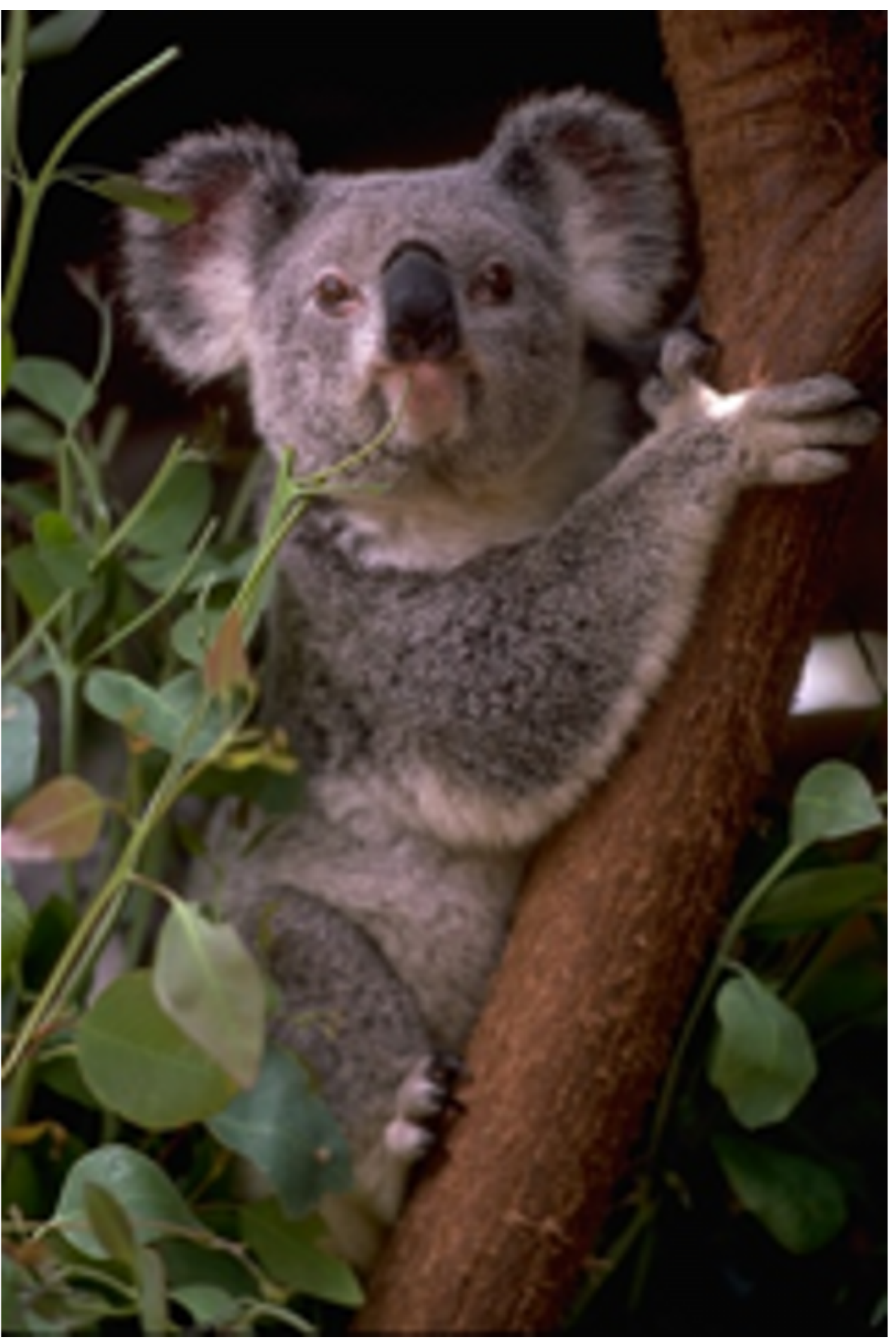} &
   \includegraphics[width=1.5in]{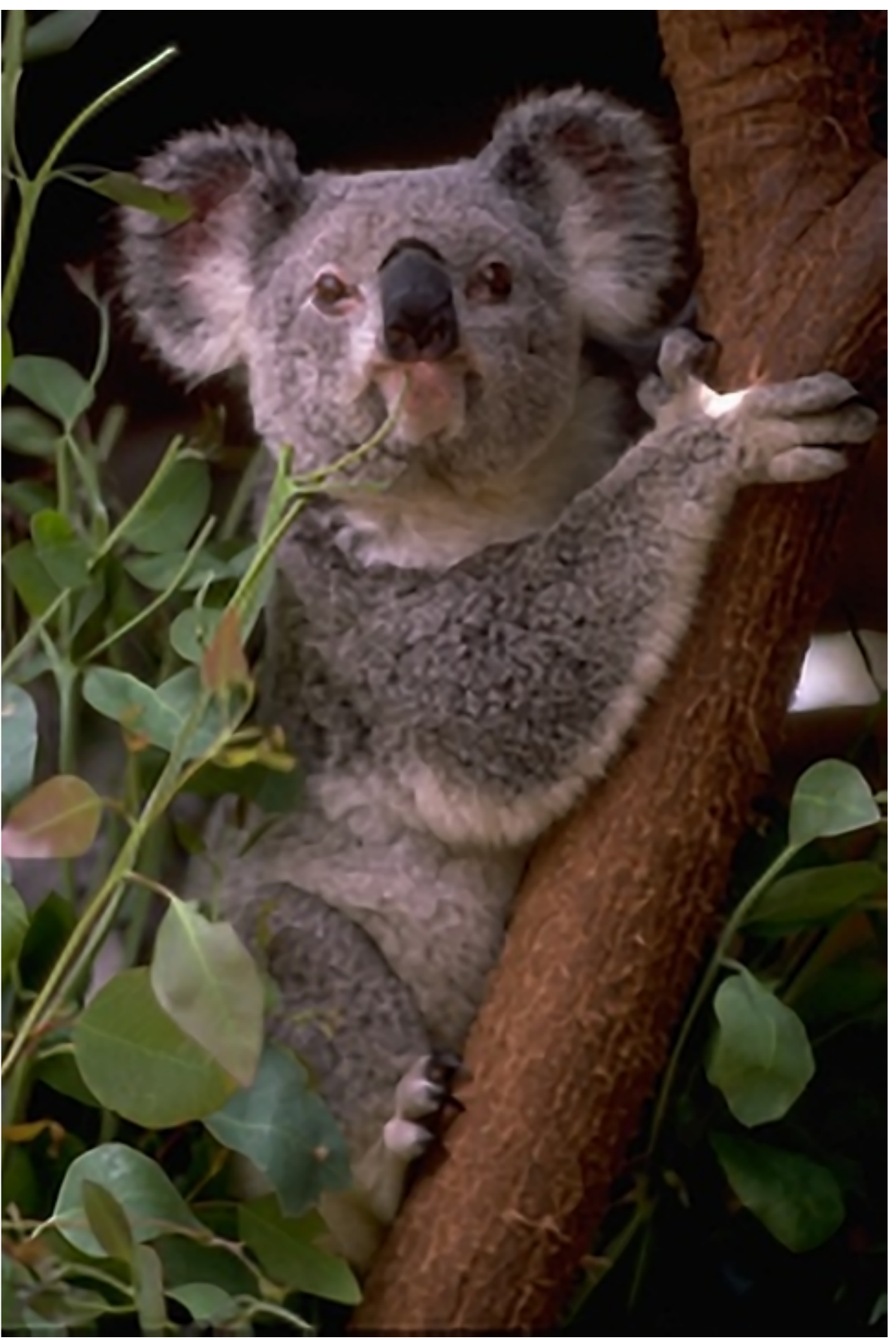} &
   \includegraphics[width=1.5in]{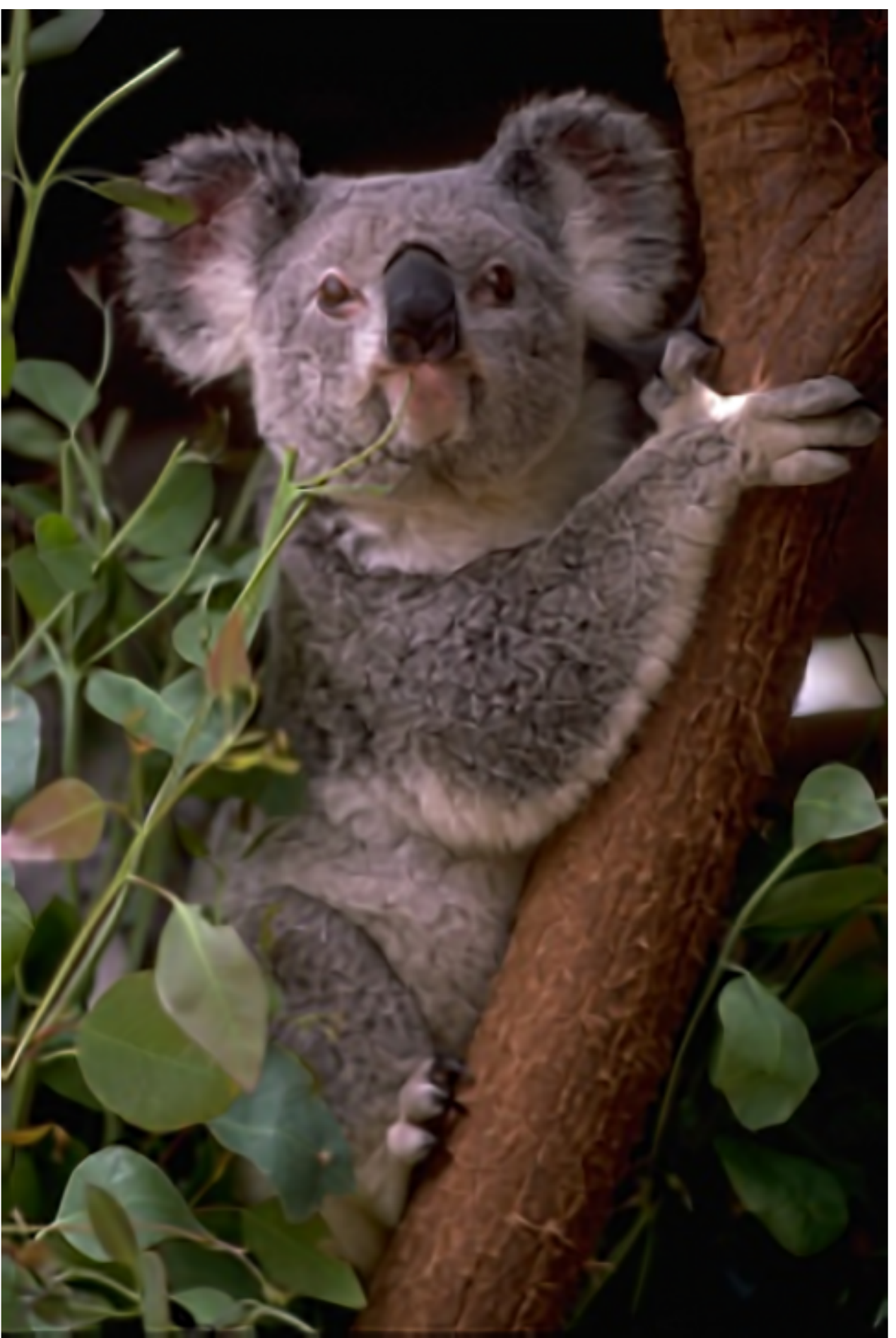} &
   \includegraphics[width=1.5in]{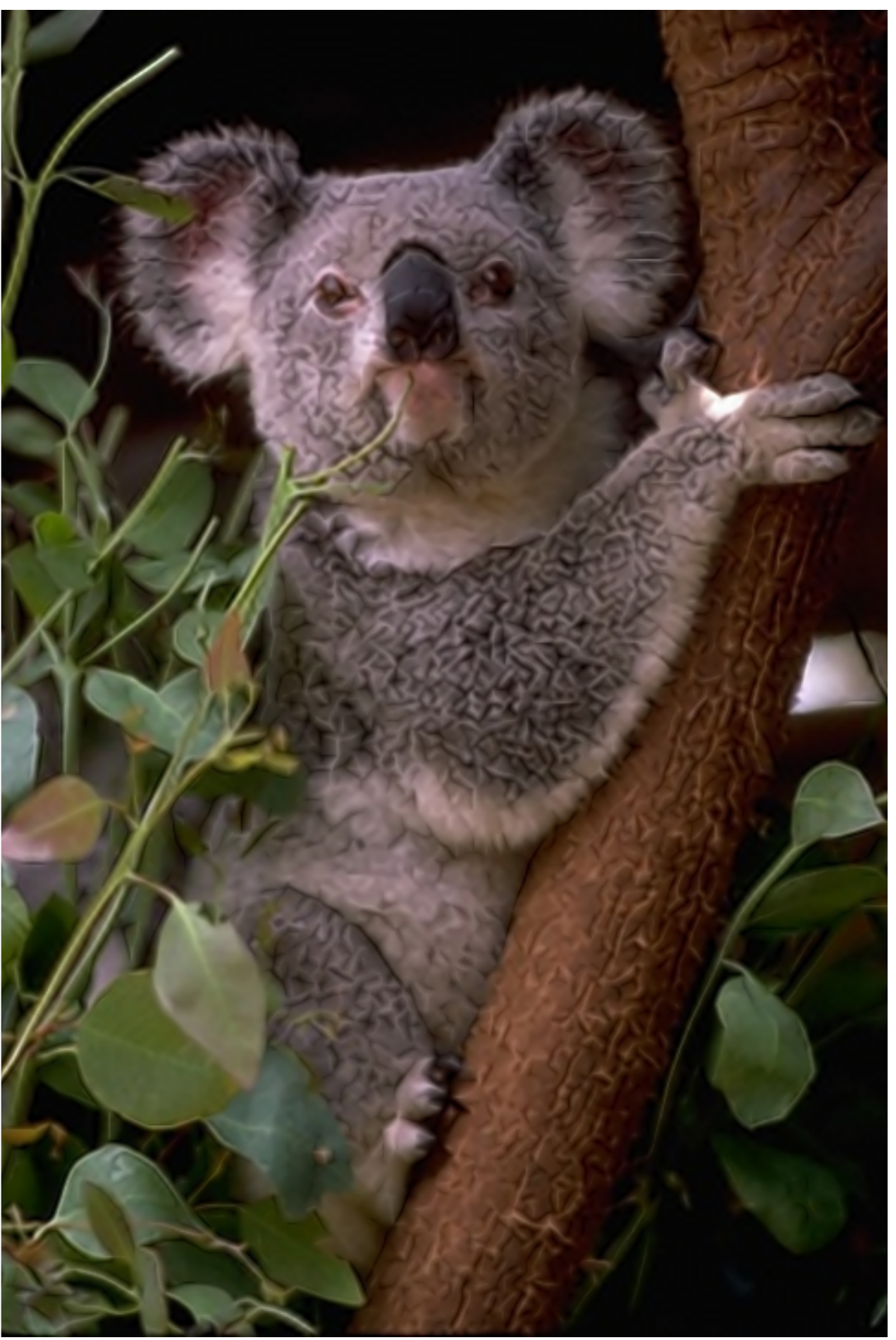} \\

   \includegraphics[width=1.5in]{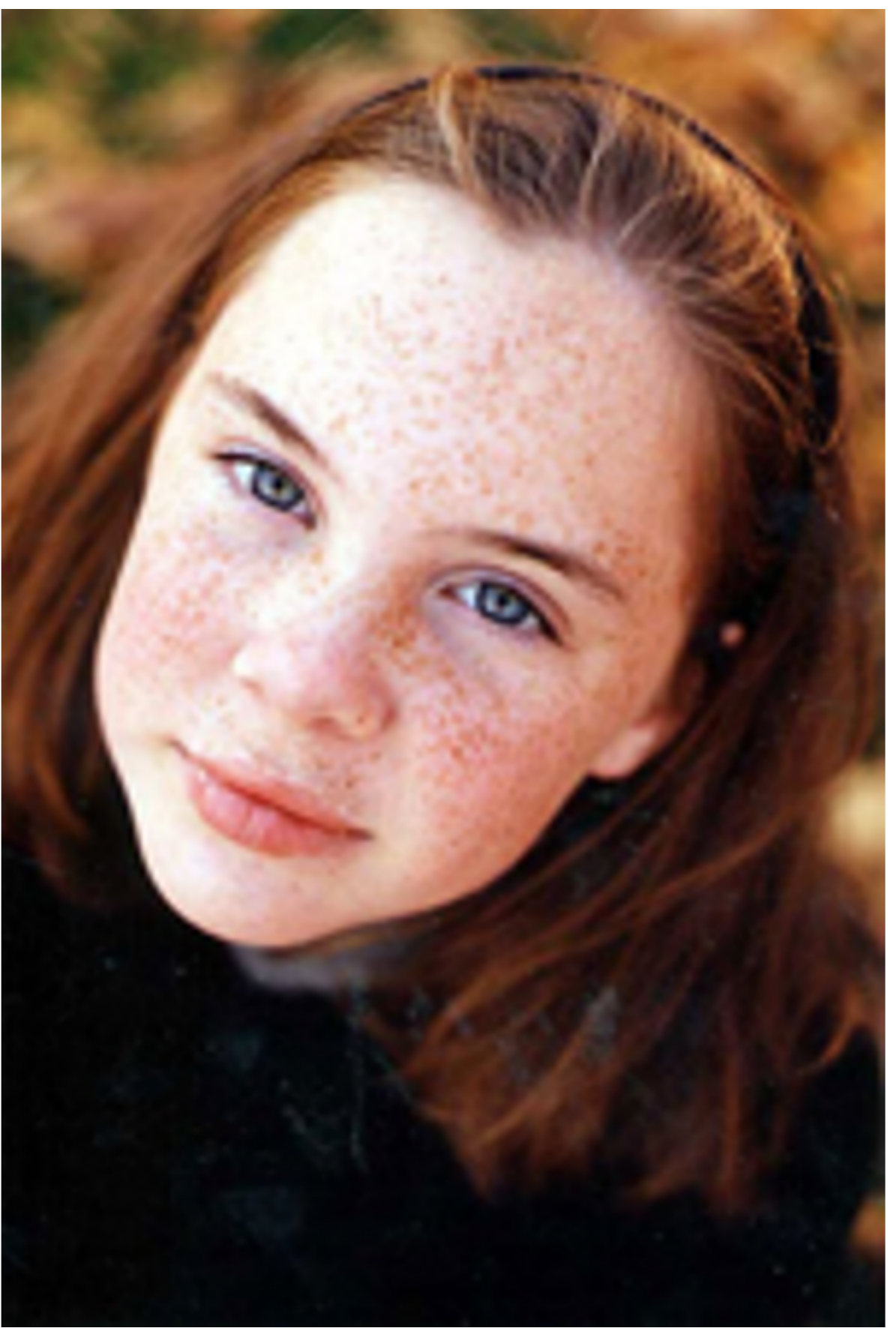} &
   \includegraphics[width=1.5in]{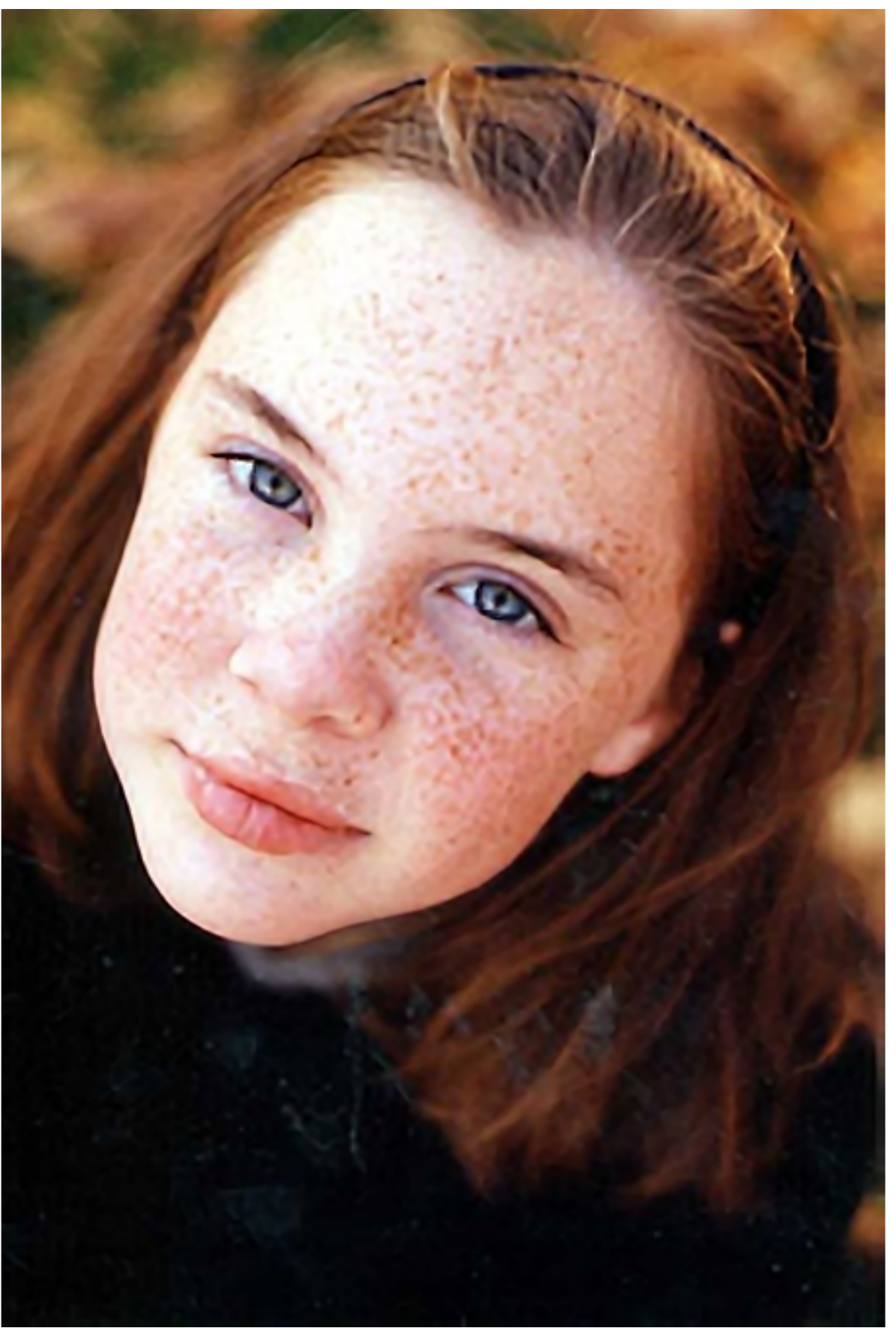} &
   \includegraphics[width=1.5in]{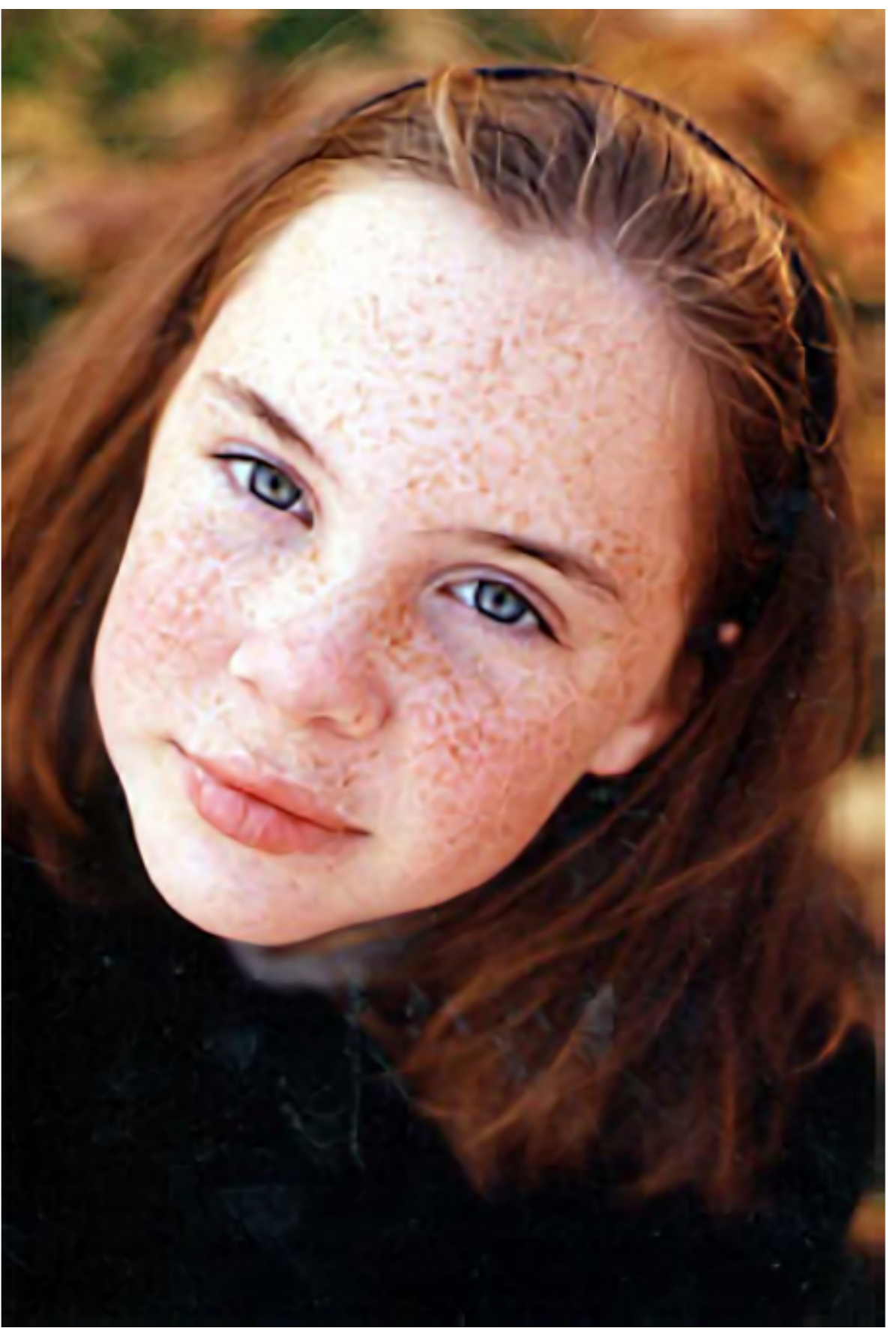} &
   \includegraphics[width=1.5in]{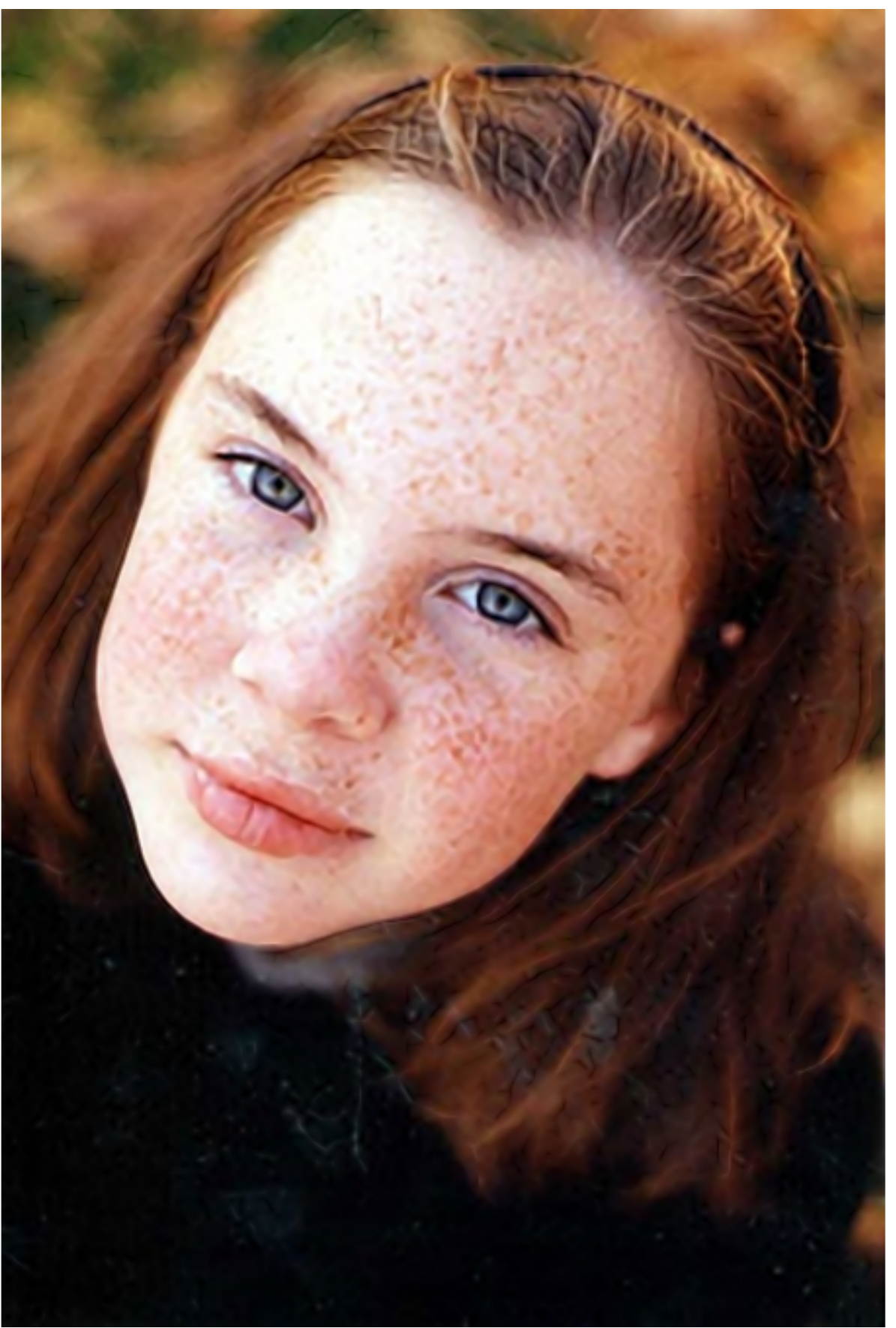} \\
{\small Bicubic interpolation} & {\small Scale-space search~\cite{irani}} & {\small Filter-based Upsampling~\cite{fattal11}} & {\small Proposed layer-based method}
\end{tabular}
   \caption{Example $3\times$ upsampling results.}
\label{fig:girl}
\end{figure*}

Figures~\ref{fig:chip} and \ref{fig:girl} provide additional results and comparisons with the state-of-the-art. Our method is able to reproduce fine-scale detail (like the girl's hair, the gentleman's hat,  and the koala's fur in Figures~\ref{fig:girl} and~\ref{fig:oldman}). Our method also reduce ringing artifacts around sharp edges (such as the thin white ridge in Fig.~\ref{fig:chip}). Please refer to the supplemental material for several additional comparisons and visualizations.

It is not typical in the literature to report upsampling results in terms of objective numerical comparisons; the reasons for this is that it is often hard to define an objective metric of image quality, and ground truths for the standard test images are not always available. Nevertheless, we present some numerical comparisons in Table~\ref{tab:comparisons} in terms of two popular metrics: peak SNR (in dB), and structural similarity index (SSIM)~\cite{ssim}. It is clear from Table~\ref{tab:comparisons} that our proposed method performs extremely favorably under both metrics. In particular, SSIM is widely considered to be a standard image quality measure for human perception; a higher value of SSIM indicates better performance; our layer-based method yields the highest SSIM value among all state-of-the-art upsampling algorithms.

We implemented our method in MATLAB\footnote{The nearest-neighbor step in Eq.~\ref{eq:nn} was implemented using the open-source MATLAB package OpenTSTool, which consists of a library mexed C++ functions.} and executed it on a Windows desktop machine with a 3.4GHz dual-core Intel-i7 processor.  On average, our method takes less than 1 minute to upsample an image of size $200\times200$ by a magnification factor of 4 in either dimension. Compared to other example-based upsampling and texture enhancement methods, this represents an order-of-magnitude increase.
A key reason for this improvement is that we favor local searches in our edge-enhancement step (Eq.\ \ref{eq:nn}). This greatly reduces computational demands, as well as makes our method easily parallelizable. Hence our proposed method can potentially be viable when implemented on a GPU.

\subsection{Comparisons to Prior Art}
Our method extends and refines aspects of various existing super-resolution and detail enhancement methods. Primarily, we have advocated to explicitly deal with upsampling of texture details, which has been the bane for many super-resolution algorithms. Our proposed algorithm bears resemblance to the scale-space search approach~\cite{ebrahimi,irani}. However, we recognize that the non-parametric search is both computationally demanding, as well as fails to reproduce fine textures, and thus discuss methods to alleviate these concerns. Our algorithm also can be linked to the non-stationary filter-based upsampling method~\cite{fattal11}, which also uses a non-parametric search step to upsample the sharp edges. However, the upsampling filters used in that method are often hard to compute for arbitrary scaling factors. Finally, we contrast our proposed method with the texture hallucination approach \cite{halluc}. We echo their concerns that textures are problematic in natural image upsampling; however, we propose a universal scheme for reproducing high-resolution textures that does not require manual intervention, obviates the need for a training database, and enjoys a great reduction in computational complexity.

\section{Conclusions}
\label{sec:conc}

In contrast to previous methods for image upsampling, our method explicitly recognizes the presence of various layers in natural images; edges and fine-scale details are fundamentally different, and thus need to be upsampled in a fashion unique to their structure. There are numerous avenues for improvement of our algorithm. The exact choice of the texture enhancement step $\mathcal{E}$ will likely influence the quality of synthesized details, as well as the speed of the overall algorithm. We used a simple, global detail enhancement procedure; a localized, region-specific enhancement might lead to better results. We defer this to future research.

{\small
\bibliographystyle{ieee}
\bibliography{upsample}
}

\end{document}